\documentclass[journal,onecolumn]{ieeetj}
\usepackage{cite}
\usepackage{amsmath,amssymb,amsfonts}
\usepackage{algorithmic}
\usepackage{graphicx,color}
\usepackage{textcomp}
\usepackage{xcolor}
\usepackage{siunitx}
\usepackage{subcaption}
\usepackage{caption}
\usepackage{hyperref}
\hypersetup{hidelinks=true}
\usepackage{algorithm,algorithmic}
\usepackage{acro}
\usepackage{booktabs}
\usepackage{multirow}
\DeclareAcronym{svm}{
  short = SVM,
  long  = support vector machine,
  short-indefinite = an,
  long-indefinite = a
}

\DeclareAcronym{agv}{
  short = AGV,
  long  =  autonomous ground vehicle,
  short-indefinite = an,
  long-indefinite = an
}

\DeclareAcronym{scnn}{
  short = SCNN,
  long  = sparse convolutional neural network,
  short-indefinite = an,
  long-indefinite = a
}

\DeclareAcronym{ndt-om}{
  short = NDT-OM,
  long  = normal distributions transform occupancy maps,
    short-indefinite = an,
  long-indefinite = a  
}

\DeclareAcronym{ndt-tm}{
  short = NDT-TM,
  long  = normal distributions transform traversability maps,
  short-indefinite = an,
  long-indefinite = a  
}
\DeclareAcronym{ndt}{
  short = NDT,
  long  = normal distributions transform,
  short-indefinite = an,
  long-indefinite = a    
}

\DeclareAcronym{ohm}{
  short = OHM,
  long  = occupancy homogeneous map,
  short-indefinite = an,
  long-indefinite = an  
}

\DeclareAcronym{ftm}{
  short = FTM,
  long  = forest traversability mapping,
  short-indefinite = an,
  long-indefinite = a  
}

\DeclareAcronym{TE}{
  short = TE,
  long  = traversability estimation,
  short-indefinite = a,
  long-indefinite = a  
}

\DeclareAcronym{mcc}{
  short = MCC,
  long  = Matthews correlation coefficient,
  short-indefinite = an,
  long-indefinite = a  
}

\DeclareAcronym{lfe}{
  short = LfE,
  long  = learning from experience,
  short-indefinite = an,
  long-indefinite = a  
}

\DeclareAcronym{hl}{
  short = HL,
  long  = hand-labelled,
  short-indefinite = an,
  long-indefinite = a  
}

\DeclareAcronym{vd}{
  short = CVD,
  long  = column vegetation density,
  short-indefinite = a,
  long-indefinite = a  
}

\DeclareAcronym{focc}{
  short = $f_{OCC}$,
  long  = column vegetation density,
  short-indefinite = a,
  long-indefinite = a  
}

\DeclareAcronym{SLAM}{
  short = SLAM,
  long  = Simultaneous Localisation and Mapping,
  short-indefinite = a,
  long-indefinite = a  
}

\DeclareAcronym{dtr}{
  short = DTR,
  long  = Dynamic Tracked Robot,
  short-indefinite = a,
  long-indefinite = a  
}

\DeclareAcronym{atr}{
  short = ATR,
  long  = All Terrain Robot,
  short-indefinite = a,
  long-indefinite = a  
}

\DeclareAcronym{ograph}{
  short = OGraph,
  long  = Online Data Graph,
  short-indefinite = an,
  long-indefinite = an  
}

\DeclareAcronym{shm}{
  short = SHM,
  long  = single head model,
  short-indefinite = an,
  long-indefinite = a  
}

\DeclareAcronym{thm}{
  short = THM,
  long  = two headed model,
  short-indefinite = a,
  long-indefinite = a  
}

\DeclareAcronym{lfe_chl}{
  short = $LfE + HL_{conf}$,
  long  = Learning from Experience with high confident hand-labels,
  short-indefinite = a,
  long-indefinite = a  
}

\DeclareAcronym{lfe_hl}{
  short = $ LfE + HL$,
  long  = Learning from Experience with hand-labels,
  short-indefinite = a,
  long-indefinite = a  
}

\DeclareAcronym{tr}{
  short = $TR$,
  long  = traversable,
  short-indefinite = a,
  long-indefinite = a  
}

\DeclareAcronym{nt}{
  short = $NT$,
  long  = non-traversable,
  short-indefinite = a,
  long-indefinite = a  
}

\DeclareAcronym{tsdf}{
  short = TSDF,
  long  = Truncated Signed Distance Field,
  short-indefinite = a,
  long-indefinite = a  
}

\DeclareAcronym{sdf}{
  short = SDF,
  long  = Signed Distance Field,
  short-indefinite = a,
  long-indefinite = a  
}

\DeclareAcronym{nerf}{
  short = NeRF,
  long  = Neural Radiance Fields,
  short-indefinite = a,
  long-indefinite = a  
}

\DeclareAcronym{gsplat}{
  short = GSplat,
  long  = Gaussian Splatting,
  short-indefinite = a,
  long-indefinite = a  
}

\DeclareAcronym{esdf}{
  short = ESDF,
  long  =  Euclidean Signed Distance Field,
  short-indefinite = an,
  long-indefinite = an  
}

\DeclareAcronym{gpu}{
  short = GPU,
  long  =  Grafical Compute Unit,
  short-indefinite = a,
  long-indefinite = a  
}

\DeclareAcronym{covar}{
  short = CoVaR,
  long  =  Conditional Value at Risk,
  short-indefinite = a,
  long-indefinite = a  
}

\newcommand{\figref}[1]{Figure~\ref{#1}}

\definecolor{britishracinggreen}{rgb}{0.0, 0.26, 0.15}

\def\BibTeX{{\rm B\kern-.05em{\sc i\kern-.025em b}\kern-.08em
    T\kern-.1667em\lower.7ex\hbox{E}\kern-.125emX}}
\AtBeginDocument{\definecolor{tmlcncolor}{cmyk}{0.93,0.59,0.15,0.02}\definecolor{NavyBlue}{RGB}{0,86,125}}

\def\OJlogo{\vspace{-4pt}$<$Society logo(s) and publication title will appear here.$>$}
\def\seclogo{\vspace{10pt}$<$Society logo(s) and publication title will appear here.$>$}

\def\authorrefmark#1{\ensuremath{^{\textbf{#1}}}}

\begin{document}
\receiveddate{XX Month, XXXX}
\reviseddate{XX Month, XXXX}
\accepteddate{XX Month, XXXX}
\publisheddate{XX Month, XXXX}
\currentdate{XX Month, XXXX}
\doiinfo{XXXX.2022.1234567}

\markboth{}{Fabio A. Ruetz. {et al.}}

\title{Online Adaptive Traversability Estimation through Interaction for Unstructured, Densely Vegetated Environments}

\author{
Fabio A. Ruetz\authorrefmark{1,2}, 
Nicholas Lawrance\authorrefmark{2},
Emili Hern\'andez\authorrefmark{3},
Paulo V. K. Borges\authorrefmark{4},
and Thierry Peynot\authorrefmark{1}}
\affil{QUT Centre for Robotics, Queensland University of Technology (QUT), Brisbane Qld 4000, Australia}
\affil{CSIRO Robotics, Data61, Pullenvale, Qld 4069, Australia.}
\affil{Emesent, Milton, Qld 4064, Australia}
\affil{Orica, Windsor, Qld 4030, Australia}
\corresp{Corresponding author: Fabio A. Ruetz (email: fabio.ruetz@gmail.com).}
\authornote{This work was supported by QUT, CSIRO, Emesent and the SILVANUS Project through European Commission Funding on the Horizon 2020 call number H2020-LC-GD-2020, Grant Agreement number 101037247. F.R. and T.P. acknowledge continued support from the Queensland University of Technology (QUT) through the Centre for Robotics.}

\begin{abstract}
Navigating densely vegetated environments poses significant challenges for autonomous ground vehicles. Learning-based systems typically use prior and in-situ data to predict terrain traversability but often degrade in performance when encountering out-of-distribution elements caused by rapid environmental changes or novel conditions. This paper presents a novel, lidar-only, online adaptive traversability estimation (TE) method that trains a model directly on the robot using self-supervised data collected through robot-environment interaction. The proposed approach utilises a probabilistic 3D voxel representation to integrate lidar measurements and robot experience, creating a salient environmental model. To ensure computational efficiency, a sparse graph-based representation is employed to update temporarily evolving voxel distributions. Extensive experiments with an unmanned ground vehicle in natural terrain demonstrate that the system adapts to complex environments with as little as 8 minutes of operational data, achieving a Matthews Correlation Coefficient (MCC) score of 0.63 and enabling safe navigation in densely vegetated environments. This work examines different training strategies for voxel-based TE methods and offers recommendations for training strategies to improve adaptability. The proposed method is validated on a robotic platform with limited computational resources (25W GPU), achieving accuracy comparable to offline-trained models while maintaining reliable performance across varied environments.
\end{abstract}



\maketitle

\section{INTRODUCTION}
Learning-based systems enable autonomous navigation in complex structured and unstructured environments where classical methods have strugled~\cite{frey2023fast, guastella2021learning}, for example, in densely vegetated environments. However, methods relying on modern learning techniques can suffer from decreasing performance when encountering previously unseen elements in the environment, resulting in unsafe behaviour or catastrophic failure. 
Many modern approaches rely on supervised data to adapt or finetune learnt models to a novel environment, to a different robotic platform or sensor configuration. This commonly requires the collection and hand-labelling of data, which is expensive and time-consuming. Hence, adapting a system to every new environment can be prohibitively expensive in terms of both time and cost. For industry applications, this presents a severe practical barrier to adopting and deploying autonomous systems in various scenarios.
Consequently, the ability of a pre-trained model to adapt to a new domain or, in extreme cases, the ability to train a new model in a self-supervised fashion directly in a novel environment is highly desirable. It may even be necessary to enable widespread deployment and adaptation of autonomous systems. Further, depending on the application, time constraints can be critical. For example, an autonomous agent intended to fight bushfires would not have time to be re-trained for each deployment location. The key factors here are the time and effort to deploy in a novel environment.

The issue of high variation and change is particularly relevant for densely vegetated environments. These environments can exhibit significant and rapid changes due to weather,  rapid plant growth, seasons or day/night cycle.  Additionally, there is a considerable variation within the flora depending on the location and the macro-climate, e.g. sun vs shade side of a hill. In essence, there can be a high amount of variability and sudden changes in the environment on a small spatial and temporal scale, as was also reported in field experiments~\cite{bradley2015scene, frey2023fast}. Hence, the ability to adapt to the environment is essential to enable robust deployment of autonomous agents in natural environments.

The recent literature on online adaptation for robotic navigation in unstructured and complex environments relies on image modalities. Images are typically labelled using self-supervised approaches, where experience (e.g. traversal, realised speed vs. actual speed, IMU signals) is associated with the image~\cite{frey2023fast, bae2023self, wellhausen2019where, kahn2020badgr}. 
The robot's experience of traversing a patch or a volume of terrain is used to associate proprioceptive sensor readings and robot state with exteroceptive sensor measurements or spatial representation. This fusion allows for the generation of self-supervised labelled data with a traversability metric or surrogate. More recent work includes geometric measurements, such as point clouds or height maps, to assist in the projection or as an additional feature for a neural network~\cite{li2023seeing, frey2024roadrunner}. However, the mentioned method solely relies on the image modality to discriminate the environment, which is prone to catastrophic failures to environmental conditions, e.g. light changes or inability to navigate due to occlusions.
Commonly, non-traversable examples are not captured during data collection to avoid the risk of damaging the robot. Instead, surrogate metrics are defined for non-traversable elements. This has been shown to lead to desirable behaviours or navigation strategies that may implicitly emerge~\cite{frey2023fast}, but may not capture the actual traversability capability of the robotic system. These methods are similar to self-supervised learning approaches that use post-processed data but require additional considerations to be deployed online.


This work proposes a novel online adaptive TE method that allows the robot to adapt to previously-unseen densely vegetated environments and enables safe navigation. The method can be trained in real-time, in situ, on the platform, with the help of a human operator guiding the robot to experience the environment. The proposed method leverages 3D probabilistic voxel maps, which are robust against single erroneous sensor readings and occlusions. Further,  an incremental data fusion scheme combines lidar measurements and robot experience into a graph-based data structure. This approach allows to retain and update temporarily evolving voxel distribution whilst minimising the computational load. This method differentiates itself from other recent works in this area, as it is a pure lidar-based geometric method achieving online \ac{TE} in vegetated environments. It provides insights when dealing with multi-distribution probabilistic voxel-based maps in the online learning domain, which have not been presented anywhere else. Further, the proposed method does not require an ``expert operator''  with significant technical knowledge to adapt a robotic system to a novel environment.

The proposed online adaptive forest traversability estimation allows the operator to quickly adapt an existing model to a new environment. The contributions of this paper are as follows: 
\begin{itemize}
 \item we present a novel online adaptive traversability estimation method that can adapt to novel/unseen environments during deployment,
 \item we propose a data graph strategy to capture and maintain a graph of the environment in a sparse form, allowing for the generation of high-quality data whilst the robot operates,
 \item we demonstrate that online adaptation can be achieved solely with self-supervised data whilst running on a robot in real-time and validate this with a model trained online for point-to-point navigation,
 \item we provide a detailed evaluation of \ac{TE} model performance based on a variety of training or adaptation schemes and network architectures, resulting in recommendations of strategies that should be used depending on the constraints,
 \item we present an extensive field test using point-to-point navigation for six different methods and models trained with different data and approaches, allowing us to gather further insight into the benefits and limitations of this approach. 
\end{itemize}

This work differs from other contributions in the field by using only lidar and a temporally evolving 3D probabilistic voxel map rather than imaging modalities~\cite{frey2023fast, yoon2024adaptive}. Further, this method allows the robot to experience the failure state of learning actual terrain traversability rather than a surrogate measure and generate accurate \ac{lfe} data. In addition, this work releases open-source codebases and additional videos on the project website~\footnote{ \href{https://fabioruetz.github.io/foresttrav_aol.github.io}{Project Website: https://fabioruetz.github.io/foresttrav\_aol.github.io}} ~\footnote{ \href{https://github.com/csiro-robotics/foresttrav}{Github: https://github.com/csiro-robotics/foresttrav }}.

\section{RELATED WORK}
\label{sec:related_work}

Traversability estimation is the assessment of a patch of terrain to determine whether a robot can enter, reside and exit this patch without entering a failure state~\cite{papadakis2013terrain}. In the literature, traversability estimation has classically been separated into geometric, appearance-based and mixed methods. Geometric methods aim to assess traversability using geometric environmental representations commonly generated from range measurements, and these methods have shown great success in rigid, complex, unstructured environments~\cite{HudTal21, tranzatto2022cerberus}. However, these fail in environments with pliable elements, such as vegetated environments, due to the inherent assumption of a rigid environment and the inability to interact with pliable elements if required to succeed in a task~\cite{frey2023fast}.  

For~\ac{TE} in vegetated environments, the use of image modalities has been considered essential by many authors to discriminate and assess complex environments, either via purely image-based methods~\cite{wellington2004online, frey2023fast} or mixed methods~\cite{frey2024roadrunner}. Pure image- or appearance-based methods rely on a single image or short sequence and can assess complex environments but are prone to catastrophic failures due to occlusion or adversarial environmental conditions, e.g. lighting conditions. Mixed methods leverage both domains, where the geometric representations are commonly used to fuse the sequential appearance-based~\ac{TE} estimates, the geometric representations are used as data structures, and the assessment relies solely on image modalities~\cite{frey2023fast, chen2023rspmp}.

General late-fusion methods combine different smaller assessments into a unified cost or risk metric, which has been shown to be a versatile general tool for assessing different traversability aspects~\cite{quigley2009ros, frey2024roadrunner, fan2021step}. Classical heuristic costmap fusion~\cite{quigley2009ros} has been replaced by~\ac{covar} risk assessment, where a general notion of risk is generated. It allows one to quantify how ``bad'' a failure case would be~\cite{fan2021step}. Recent work has performed 2.5D~\ac{TE} estimation for high-speed, off-road vehicles in unstructured, grassy environments~\cite{frey2024roadrunner} in large 2.5D maps. However, this assessment of vegetation remains in the image domain and does not consider full 3D. Separation in support surface (ground) and pliable elements has been proposed with different variations~\cite{wellington2004online, bradley2015scene, li2023seeing}, but they rely on the image modality as the discriminator.

Traversability estimation for vegetated environments was explored by~\cite{ahtiainen2017normal} with the~\ac{ndt-tm} representation capturing salient information of the environment at large voxel resolutions (0.4~\si{m}). Ruetz et al.~\cite{ruetz2022FTM} introduced additional features and distributions for smaller voxel representations and a real-time capable system~\cite{ruetz2024foresttrav}. Our prior work ForestTrav, currently the only lidar-based method using 3D voxel representations, demonstrated that geometry, salient voxel features, and context are sufficient for a geometric representation to assess complex environments in real-time~\cite{ruetz2024foresttrav} and allow for save navigation. The work relied on \ac{scnn}, a deep learning method that can implicitly represent missing data, to allow for rapid learning and inference in 3D.  The work fused~\ac{lfe}, self-supervised data from the robot experiencing the environment, and hand-labelled (HL) data to generate a high-quality, combined data set called ~\ac{lfe_hl} data. 

Online or adaptive traversability estimation methods aim to adapt a model to novel environments online on the running robot in the field~\cite{wellington2004online, frey2023fast, kim2006traversability, hadsell2009learning, kahn2020badgr}. They leverage similar principles as self-supervised learning approaches for~\ac{TE} but require the generation of additional training data and adapting the model on the robot in the field. These approaches require a deep understanding of the problem and the robotic system, as well as a computationally efficient implementation to be able to achieve this.

Early works by Kim et al.~\cite{kim2006traversability} demonstrated the use of image modalities to train an online adaptive visual \ac{TE} method. They associated image features with self-supervised labels gathered through interaction using a clustering algorithm. They used classical appearance colour/texton information and contact sensors to associate image and robot experience in a grid representation. Hadsel et al. ~\cite{hadsell2009learning} proposed to train a multi-class classifier online using short-distance measurements and self-supervision as a source for the ground-truth labels. The rule-based classifier was trained online, but the CNN feature generator was trained offline on images of different environments. 

More recent work has leveraged the rapid advancements in deep learning frameworks and hardware in recent years~\cite{kahn2020badgr,frey2023fast,yoon2024adaptive}. 
Kahn et al. ~\cite{kahn2020badgr} demonstrated an end-to-end navigation method aimed at outdoor environments. They formulated the problem of the end-to-end method using an action-predictive deep neural network, which, given a set of sequential actions, predicts the following K sequential events. They used off-policy self-labelled data to collect data and allow the system to fail. The resulting method was able to navigate successfully using only images in various outdoor scenarios.

Frey et al. presented an online adaptation method in~\cite{frey2023fast} that relies on self-supervised using images and differences of estimated and realised velocities. A mission graph allows tracking poses and traversability-related signals (velocity differences) and can associate this with features from image segments. These are generated using DINOv2 on image segments generated by the super-pixels algorithm. The CNN estimating the traversability was trained incrementally using batches online, allowing the model to learn in minutes. The pixel-wise \ac{TE} score is projected onto a 2.5D height map used by the local planner. 
The learning objective is formulated as a positive-unknown approach. Self-supervised and labelled data are used as positive examples, and the unlabelled data are used as negative examples, with an additional confidence loss to enforce a compact feature embedding for the positive (traversable class) and a clear separation to the unknown class. The learned traversability is based on the similarity of previously traversed areas and the model's confidence in the labelled and unlabelled data. The resulting behaviours allowed the robot to navigate safely in areas it is confident it has seen before. The natural behaviour that emerges shows the robot prefers an easier path, learning to walk on a gravel path rather than the grass beside it. Further, this method has been demonstrated in various natural, unstructured environments. The authors noted that the models do not generalise well to nearby environments. Compared to the previous method, this method relies on an intermediate geometric representation and classical planner to navigate. 

Yoon et al. ~\cite{yoon2024adaptive} provided an online adaptive learning approach based on a 2D grid-based lidar elevation map for off-road environments. It aims to filter out patches of vegetation for a gravel-based road. Their adaptive replay approach allows for maintaining a data set with the relevant information. Additionally, the authors claim to estimate the aleatoric uncertainty while estimating the traversability of local grid map patches. They demonstrate their work on an off-road gravel path with different densities of vegetation between the tracks.

The work presented in this paper differs as it presents an online adaptive learning method for a probabilistic voxel representation. Whilst the community has generated a large amount of understanding for universal feature generation and online learning on images, the space of 3D probabilistic representation has not been well explored. A key critical difference between using 3D-voxel distributions and images as network input is that the voxel distributions evolve over time and need to be updated; they need sufficient measurements to generate a good representation of the environment of traversability through self-supervision. The concepts known from the image domain do not translate well, e.g. storing 3D maps at fixed intervals. This produces large quantities of duplicate and potentially conflicting data, e.g. when the traversability of a voxel changes due to the self-supervision. The large amount of data leads to rapidly increasing training times. This proposed method addresses these issues.

\section{METHODOLOGY}
\label{ch7:methodology}
\begin{figure*}[ht]
 \centering
 \includegraphics[width=\textwidth ]{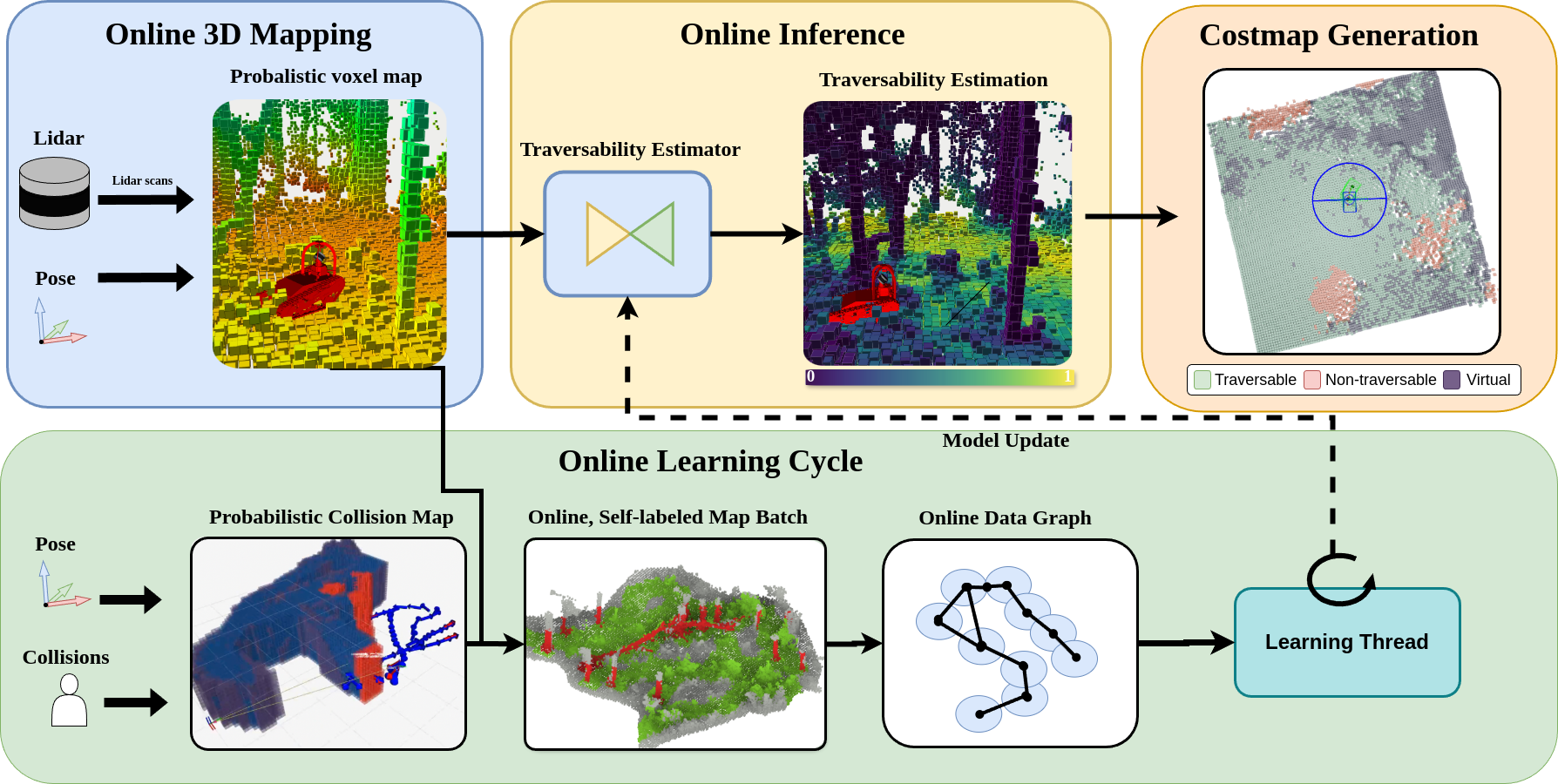}
 \caption[Online Adaptive TE Method Overview]{The overview of the online~\ac{TE} adaptation method. The \textbf{ForestTrav} method is augmented with new modules to facilitate online learning. The online node generates and trains the new model from the self-supervised labelled data collected online, stored as a graph representation. A fusion module generates the data, combining the collision map and 3D probabilistic environmental representation at fixed-time intervals into the \ac{ograph}.}
 \label{fig:odap_overview}
\end{figure*}

The method presented in this paper allows for human-assisted online~\ac{TE} to rapidly adapt to a novel environment. This is achieved through a human operator directing the robot to interact with the environment to experience safe traversal as well as collision states. The learnt traversability model adapts to the new experience data to improve its future \ac{TE} predictions in the novel environment. An overview of the method can be seen in \figref{fig:odap_overview} and shows the novel contribution that allows the method to learn and adapt online. The online 3D probabilistic mapping (blue block) is adapted from our previous work ForestTrav~\cite{ruetz2024foresttrav}, and we use a similar architecture for voxel-wise \ac{TE} but with a single neural network instead of an ensemble due to computational constraints. 

The novelty of this work is the addition of the online learning capability, allowing the TE model to adapt online using in-situ robot experience only (green block). The learning module starts a learning cycle at discrete times $t_k$ and occurs at fixed-time intervals $\delta_{t}$.  
A human operator starts the online learning process if the current~\ac{TE} model shows insufficient performance in a new environment. During online adaptation, the operator controls the robot using an RC remote and records the collisions, allowing it to capture relevant robot-environment interaction while minimising the risk of damaging the system. The collision states and associated poses are used to generate the probabilistic collision map $\mathbf{M}^C$, described in Section~\ref{subsec:collision_mapping}. The online update process is stopped after a pre-defined condition (in this implementation, we use a fixed-time interval), after which the robot can continue its mission autonomously.

The collision map at $t_k$ is fused with all 3D probabilistic maps $\mathbf{M}_{t_{k-1}:t_{k}}$ between the end of the previous learning cycle $t_{k-1}$ and the current one $t_k$. The resulting fused \ac{lfe} representation is the online, self-labelled map batch that is the current best belief of the \ac{lfe} map and is used to update a sparse graph of the environment, \ac{ograph}. Each graph node is associated with a keypose (similar to keyframes) of the trajectory traversed by the robot. The edges maintain the ordered poses of the trajectory and help with the rapid traversal of the graph. Each node contains a locally-fused~\ac{lfe} map, which is used as a training sample. The graph serves as a sparse solution for the fused~\ac{lfe} map since maintaining a global map is not computationally feasible. It still allows us to update and maintain the temporally evolving collision states and distribution of the 3D map of the environment that the robot experienced.

Finally, \ac{ograph} is used by an online training module that can either train or finetune the model online using the submap of the nodes as training samples. The method trains the model for a small fixed number of training epochs ($n_{adapt}$) to ensure learning finishes before the next adaptation cycle starts. Once a model is trained, the weights of the \ac{scnn} of the traversability estimation module are updated with the newly trained one.

The last block (\figref{fig:odap_overview}) captures the generation of the conversion of the 3D~\ac{TE} map to a valid 2D costmap. This block estimates the ground, associates a traversability cost for each ground cell and generates estimates for unseen cells (virtual costs). The resulting costmap is then passed to a planner.

In this paper, we examine two different scenarios in the online-learning context. In the first scenario, a pre-trained model is available, but the training data is not. We assume the pre-trained model has been trained on high-quality data; in the case of this paper, we use~\ac{lfe_hl} data. In the second scenario, a randomly initialised (untrained) model is trained from scratch on the online data set. 

A key critical difference between using 3D-voxel distributions and images as network input is that the voxel distributions evolve over time and need to be updated. They need sufficient measurements to generate a good representation of the environment and of traversability through self-supervision. 

\subsection{Problem Definition}
\label{subsec:problem_def}
This work assesses the traversability of unstructured, natural, vegetated environments in full 3D to allow for the autonomous operation of \iac{agv}. Each voxel $m \in \textbf{M}$ of the voxel map $\mathbf{M}$ is assumed to have a true binary, exclusive traversability state $\tau \in \{ TR, NTR\}$ where $m_\tau = TR$ means that the voxel $m$ is traversable (the positive case) and $m_\tau = NTR$ means that cell $m$ is non-traversable (negative case). To estimate this traversability state, we store a vector of probabilistic features in each voxel, with features estimated from lidar observations. We use a learning approach that learns a parametric function $ p_\tau (\tau=TR | M) = f_\Theta(x): \mathbb{R}^n \rightarrow \left[0, 1\right]$, where $\Theta$ are function parameters, $x \in \mathbb{R}^n$ are the input features and $p_\tau$ is the belief that the voxel is traversable given the state of the 3D probabilistic voxel map. 
 
\subsection{3D Proablistic Representation}
\label{subsec:map_rep}
In this work, we use a 3D probabilistic voxel map $\mathbf{M}$ to represent the environment. The map $\mathbf{M}$ is a tessellation of non-overlapping 3D cubes called voxels $m$. Each voxel stores and maintains a probabilistic estimate of its state from a continuous stream of observations, referred to as voxel distributions. Voxel distributions are assumed independent of the adjacent voxels. These distributions are generated and updated continuously with the incoming stream of sensor data, making them time-dependent. This work considers additional distributions beyond the commonly-used occupancy~\cite{hornung2013octomap}, and the different tracked distributions are denoted with a suffix $X$ for a single voxel, i.e. $m_X$. The different distributions are commonly referred to as layers of the map $\mathbf{M}$, i.e. the layers that capture a particular property. For example, the occupancy layer of the map $ \mathbf{M}_{OCC} \subset \mathbf{M}$ contains only the occupancy distribution of the voxel $m_{OCC} \in \mathbf{M}_{OCC}$.


\subsubsection{Voxel Distributions and Features}
\label{subsubsec:voxel_dist}
In this work, we track the following mix of distributions and features shown below: 

\begin{equation}
    \label{eq:dist}
      \begin{array}{ll}
        \textrm{NDT-OM dist:} & m_{OCC} = [p_{OCC}, N_{OCC}, \mu_{NDT}, \Sigma_{NDT}] \\
        \textrm{NTD-TM dist:} & m_{ndt-tm} = [N_{hit}, N_{miss}] \\
        \textrm{Intensity dist:} & m_{I} = [\mu_I ,\sigma_I] \\
        \textrm{Muti-return: } & m_{MR} = [N_{MR} ] \\
    \end{array} 
\end{equation}

At the core of the representation is the \ac{ndt-om}~\cite{saarinen2013normal} representation $m_{OCC}$. For each voxel, the occupancy $p_{OCC}$ is associated with a 3D spatial Gaussian distribution (parameterised by the mean $\mu_{NDT}$ and covariance $\Sigma_{NDT}$) of the endpoint locations of lidar observations ending in that voxel. This differs in two critical ways from the commonly used binary occupancy representation. Firstly, the 3D Gaussian can be used to represent elements at sub-voxel resolution and estimate more complex geometry. Secondly, this representation does not assume that the whole voxel is occupied but rather associates the occupancy with the 3D Gaussian distribution. By considering the proximity of a ray to the distribution of a voxel, the resulting representation is more resilient to erosions of thin elements. 

From~\ac{ndt-tm}~\cite{ahtiainen2017normal}, we leverage the notion of permeability, the likelihood of a ray passing through a distribution using the statistical number of hits and misses $N_{hit}$ and $N_{miss}$. The permeability describes how the chance of a ray passing through a voxel due to many small or scattered elements. This frequently occurs in nature, e.g. patches of leaves or grass blades, and has shown to be a salient descripor~\cite{ahtiainen2017normal}. Further, the intensity mean and variance of the laser returns have been shown to help differentiate chlorophyll-rich vegetation from other elements in the environment. In practice, the VLP-16 used has not seen a clear separation as reported in~\cite{ahtiainen2017normal}, but we have seen that this does help discriminate the environment in our prior work~\cite{ruetz2022FTM}. 

Lastly, we track the number of second returns per voxel $N_{MR}$ associated with the first return; the VLP-16 used provides two returns. We can observe second returns on elements with thin or sharp edges, e.g. leaves, thin stems or blades of grass. 

We use the parameters defining the distributions directly as features, thereby avoiding any feature computation. E.g. the inclination from the Eigen Value decomposition the covariant matrix. 

\subsubsection{Self-Labelling Through Collision Mapping}
\label{subsec:collision_mapping}
This work leverages robot experience by combining the observed collision states $c_t = \{TR, NTR\}$ with the robot state $s_t \in \text{SE}(3)$ (3D position and orientation) for all times $t$ and fusing them into a collision map using a stationary Bayesian filter. The formulation is similar to that used in occupancy mapping, but the robot itself is treated as a sensor. The likelihood of a voxel being traversable or collision-free is given by the likelihood $p_{m_\tau} = p( m_\tau| c_{1:t}, s_{1:t})$, and the corresponding log-likelihood formulation is:
\begin{equation}
    \label{eq:lodds_cm}    
    l(m_{\tau} | c_{1:t}, s_{1:t}) = l(m_\tau | c_t,  s_{t}) +l(m_{\tau} | c_{1:t-1}, s_{1:t-1})
\end{equation}

\begin{figure}[ht]
    \centering
    \includegraphics[width=0.7\columnwidth]{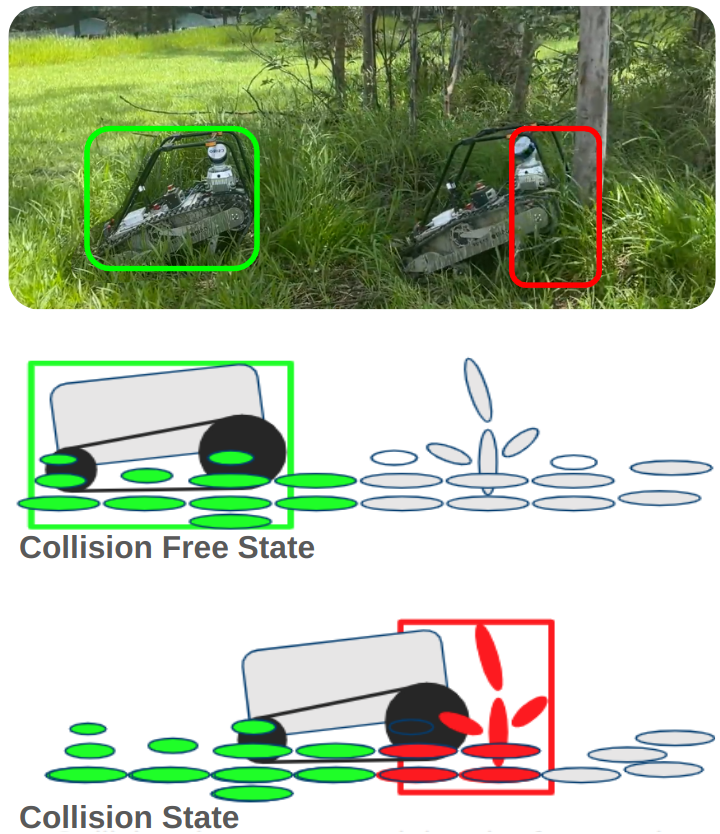}
    \caption{Visualisation of the Learning from Experience of the robot. The top image shows the robot in a real-world scenario, with green for the collision-free state and red for the collision state. The middle image shows the robot labelling the 3D probabilistic map as collision-free, ellipsoid representing the NDT representation of a voxel. The bottom image shows the robot in collision, labelling the voxels as non-traversable in the bound box at the front of the robot, with the bounding volume extending a short distance beyond the robot chassis.}
    \label{fig:robot_collisions}
\end{figure}

The first term $l(m_\tau | c_t,  s_{t})$ is the ``inverse sensor model''. A bounding box is generated for each robot state $s_t$, and all voxels intersecting with the bounding box are updated with the associated collision state $c_t$ at time $t$ with a fixed probability, see \figref{fig:robot_collisions}. The subset of the voxels that intersect with the bounding box of a collision state is referred to as ``collision state voxels'' $\hat{m}_{C}$ for the current robot state $s_t$ and collision state observation $c_t$. The update probabilities given robot collision state $p_{\tau}= p(m_\tau = TR | c_t, s_{t}) $, and $p_{NTR}= p( m_\tau = NTR | c_t, s_{t})$ are either decreasing or increasing the belief that a voxel is non-traversable. In the case of collisions, the volume starts behind the front plate and extends beyond the robot chassis one to two voxels beyond the robot chassis. Specifically, in our case, it start\qty{0.1}{\m} behind the front of the robot chassis and extends 1-2 voxels beyond it, \qty{0.2}{\m}  at \qty{0.1}{\m} voxel resolution. The distance threshold values were found heuristically based on the environment and the voxel resolution. The rationale is that obstacles preventing the robot from advancing are commonly larger elements, such as tree trunks or thick, bramble bushes. Extending the bounding box beyond the robot's direct interaction point allows the method to capture the bulk or mass of the obstacles.

The voxels labelled through the robot interaction are called \ac{lfe} data. The post-processed data are generated by hand-labelled data combined with \ac{lfe} data, denoted as \ac{lfe_hl} data; see ForestTrav for additional details~\cite{ruetz2024foresttrav}. There is a difference in maps from LfE and LfE+HL data due to the trajectory errors. LfE data relies on the trajectory of the online SLAM method, which has a larger error than the post-processed trajectory used for the LfE-HL data~\cite{ramezani2022wildcat}. 
 
\subsection{Model Architecture}
\label{subsec:model_def}
The estimator and features used in this work were initially presented in ForestTrav~\cite{ruetz2024foresttrav}. In comparison, this work relies on a single model and not an ensemble due to the computational constraints of online training. The estimator architecture is a UNet encoder-decoder network with skip connections, a commonly-used architecture in many different classification or segmentation methods~\cite{uzawa2022end}. To leverage the structure and efficiency of voxel-based representations, this work uses~\acp{scnn}~\cite{tang2022torchsparse}. \Iac{scnn} implicitly represents empty or missing data for 3D representations, allowing for rapid inference and learning, and implicitly encodes context by the structure of the representation. The training loss function is the binary cross entropy loss~\cite{good1952rational} between the traversability prediction $p_\tau$ and the label $m_\tau$, summed across all labelled voxels. Specifically, this work uses a 4-layer UNet architecture with a 16-channel input layer, fully connected skip connections and convolution kernel size $k=[1,2,2,2]$, see~\cite{ruetz2024foresttrav}.

\subsection{Online Data Generation}
\label{ch7:ograph}
Maintaining a global dense voxel map is computationally infeasible due to the intractable amount of memory required. Thus, we introduce a novel data fusion approach to combine the different distributions online into a graph representation. The data is stored in a graph-based representation \ac{ograph}, where each node \textbf{n} is represented as a 3D point and is a position of the robot trajectory in the global static frame. A new node is generated from a previous robot position if its position of any other node exceeds a threshold $d_{node} = 0.5m$. Each node contains a small, sparse local voxel map with all the voxel distributions and the associated collision probabilities, not the fully dense collision map.  The local map of each node only contains information in a cylinder around the position of the node ($r < r_{max}$, $z_{min} < z < z_{max}$). The assumption is that the probabilistic voxel and collision maps will converge to a steady state with sufficient measurements. This is highly dependent on viewpoint, the distance between the sensor and the robot, and the environment in general. Under this steady-state assumption, the voxel distributions are updated with the newest measurements when available. This means the distributions within the voxel can come from different temporal maps $\mathbf{M}(t)$. The benefit of representing the environment based on nodes is to collapse many poses into a sparse representation and allow the method to update only a (comparatively) small amount of data each time. The graph is updated at fixed intervals, dictated by the time allowed between training cycles $\delta_{t}$.

\subsubsection{Map Batch Generation}
In the first step, an intermediate representation, called ``Online, Self-labelled Map Batch'' as seen in \figref{fig:odap_overview}, is generated. The figure shows the map that is generated between the training cycles $t_{k-1}$ and $t_{k}$ with $j$ discrete time steps between the two. At each time step, a snapshot of the local map around the robot and the robot states are buffered. The labels are generated through \ac{lfe}, and the same colour schemes are used as above for illustration. The grey elements are the unlabelled data, and each visualised point contains the voxel distributions. Formally, the ``Online, Self-labelled Map Batch'' is the fusion of the temporal voxel map $\mathbf{M}_{t_{k-1}:t_{k}}$, the robot states $s_{t_{k-1}:t_k}$, and the collision map at $\mathbf{M}_{t_k}^C$. The temporal fused map $\mathbf{M}_{t_{k-1}:t_k}$ is generated by stitching together the $j$ maps to generate a single large one. Each temporal slice $\mathbf{M}(t)$ contains only a small, cylindrical probabilistic map around the pose $s(t)$. The buffering of $\mathbf{M}(t)$ ensures the availability of all the maps in the interval; the robot may move further than the bounds of its dense, probabilistic robot-centric map. When combining the probabilistic maps of different times, a given voxel may have multiple distribution values that differ (as they evolved over time). In the case of multiple distributions, the newest measurements are used under the steady-state assumption for voxel distributions. 

The collision map $\mathbf{M}_{t_{k-1}:t_k}^C$ is generated as explained in Section~\ref{subsec:collision_mapping} using the probabilistic formulation. Only the collision states (poses and collision label) are stored in practice. Note that the collision map is generated using all previous collision states and robot poses, and the dense collision map is deleted as soon as it is fused with the feature map. The collision states are fused into the temporal voxel map to form the fully fused representation. This unified map is then used to update the graph.

\subsubsection{Graph Update}
The graph is updated in two sequential steps. The first step generates the new set of nodes, given the new robot states $s_{t_{k-1}:t_k}$. The states are processed in ascending order of time. As an initial step of the update, new nodes are allocated for all the poses that are a minimum distance from all other existing nodes in the graph. Newly allocated nodes are initialised, with all measurements falling into their local regions. In the second step, all existing nodes in the graph need to be updated if the data of the ``Online, Self-labelled Map Batch'' overlaps with the map of a node. For each node, a voxel-wise association of the new data is performed and updated (with the steady-state voxel assumption). Since the context or surroundings are necessary for~\ac{scnn}, we have found it beneficial to have overlapping data, where data can be in one instance on the fringe of the local map and in another instance in the centre of a local map. The radius of the cylinder and other values are guided by the robot's dimensions and the width of the cubes used for training, as proposed in ForestTrav~\cite{ruetz2024foresttrav}. Specifically, 32 voxel width at \qty{0.1}{\m} voxel resolution for a cube width of \qty{3.2}{\m} .
 
\subsection{Online Learning Module}
The online adaptation module trains or finetunes a model using incremental online data. With each update of \ac{ograph}, a ``training cycle'' starts. For each ``training cycle,'' a single model is trained for a fixed number of epochs. The estimator is then updated with the new weights. The online training module can only train one model at a time and in cycles due to the constraints of the CPU and GPU. If the training of a model takes more than the allocated time, the data of the missed training cycle is merged into the next one. 

\subsection{Costmap Generation}
\label{subsec:cotmap_generation}
\begin{figure}[ht]
 \includegraphics[width=\columnwidth]{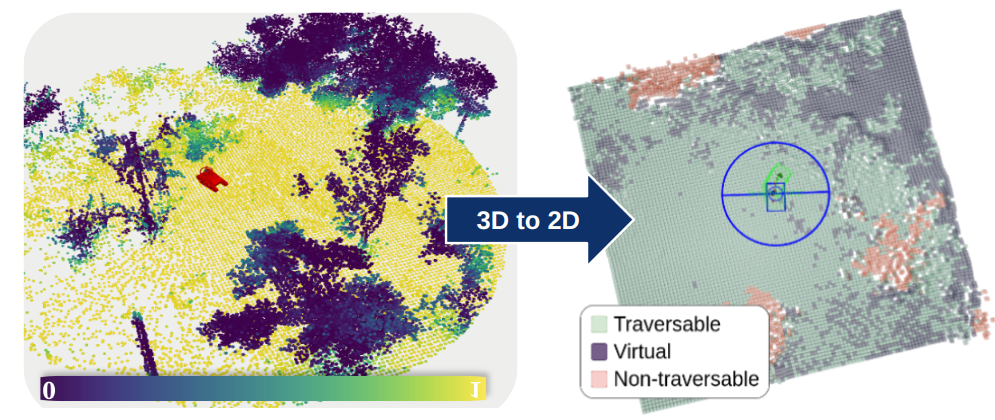}
 \caption{The Left image shows a 3D traversability map, and the right image shows a costmap with traversable (green), non-traversable (red) and virtual surfaces. The traversable elements can have low or high costs. Virtual cells are cells that have not been directly observed, and the cost is estimated based on their surroundings. }
 \label{fig:costmap_gen}
\end{figure}

The 2D costmap is generated in three steps from the 3D~\ac{TE} map: 1) support surface map, 2) virtual surface estimation, and 3) cost conversion, which are detailed below. A costmap is shown in \figref{fig:costmap_gen}. There are three colourisations: traversable, non-traversable and virtual. The green cells are traversable but also contain a cost. A traversable cell may have high costs if they are close to non-traversable elements. The virtual costs that are traversable are shown as dark blue and induce an additional cost for the planner of cells that have not been observed. The red cells are the fatal cells that the robot is not allowed to interact with. 

\subsubsection{Support Surface Generation}
For a given~\ac{TE} map, the ground or support surface is estimated per column of the map, i.e. for fixed $x$ and $y$ coordinates and variable $z$ coordinates in a gravity-aligned map. In each column of the map, the voxel with the lowest $z$ coordinate containing measurements and not higher than a heuristic limit ($z_{max}$) with respect to the robot is defined as the ground voxel. The limit $z_{max}$ is required to avoid considering overhanging elements as ground voxels, e.g. branches or tree canopy. The mean traversability probability of a cell of the costmap is the mean average of the~\ac{TE} of all the $N$ voxels above and including the ground voxel \ac{TE}. In our experiments, we use $N=10$ with voxel size \qty{0.1}{\m} to ensure only the area the robot passes through is assessed and to ignore elements far above the ground.

\subsubsection{Virtual Surface Estimation}
\label{subsec:virtual_surface_est}
In complex environments such as the dense vegetation we are tackling in this work, there can be frequent occlusions due to clutter, resulting in the support surface map containing many holes or unobserved cells.  Hence, we employ a two-stage virtual surface generation that estimates the mean traversability probability for unobserved or virtual cells from their surroundings. A two-stage approach is used. The virtual cells are generated by averaging the traversability probability and height $z$ using a $K \times K$ averaging kernel around an empty cell, given a minimum number of non-empty neighbours $N_{Adj}$. This process is run twice. In the first iteration,  the neighbours need to be non-virtual to be considered. This results in growing virtual estimations around existing measurements. The second iteration differs in that virtual surfaces are now also used to generate the estimates and will flood-fill all the gaps. Using a flood fill directly can introduce unwanted artefacts.

\subsubsection{Cost Conversion}
Finally, a valid costmap requires a cost function to transform the mean cell \ac{TE} $\bar{p}_\tau \in [0,1]$ to a cell cost $c \in [0,\inf)$. An exponential decay function is used to encourage planning in low-cost areas
\begin{equation}
 \label{eq:2d_cost}
 c = \left\{
 \begin{array}{ll}
 10.0*exp{(-6.0\cdot \bar{p}_\tau^2)} + 1.0 & \textrm{if} \quad p_\tau > \lambda_{\tau} \\
 \inf & \textrm{otherwise}
 \end{array} \right.
\end{equation}
where $c$ is the cell cost, $\bar{p}_\tau$ is the mean \ac{TE} and $\lambda_\tau$ is the traversability threshold. Fatal cells are all the cells whose cell cost is above the threshold; they are set to infinite and are visualised as red in the costmap in~\figref{fig:costmap_gen}. The exponential cost decay encourages the planner to find paths with small costs and thus take less risky paths. However, some of the green cells in the costmap can have high cell costs, close to the fatal threshold.  

The 3D to 2D costmap generation is a heuristic-based conversion. Values used for online deployment $N=10$, resulting in considering the cost map to the height of the robot with padding, $z_{max} = 0.5$, the minimum required number of neighbours is $N_{Adj} = 5$, with a $5x5$ mean averaging kernel, and the threshold of $\lambda_\tau = 0.3$. These values were determined heuristically by an expert with knowledge of the environment, the robot dimensions and observation in the field.

\subsection{Platform, Implementation Details and Learning Parameters}

\subsubsection{Platform and Sensor Suite}
To implement and validate the proposed approach, the \qty{35}{\kg} tracked platform Dynamic Tracked Robot (DTR),  shown in \figref{fig:robot_collisions}, was used with a sensory pack. The sensory pack consists of a rotating Velodyne VLP-16 and IMU. The lidar is angled at \qty{35}{\degree} and performs a complete revolution at \qty{2}{\Hz}. A detailed description of the navigation software CSIRO NavStack can be found in~\cite{HudTal21}, and the local planner we use is the hybrid A* planner~\cite{dolgov08}. The robotic platform runs an NVIDIA Jetson Orin NX (\qty{25}{\watt}, \qty{16}{\giga\byte} RAM and 100 TOPS) for the learning component of this work.

\subsubsection{Learning and Architecture Parameters}
\label{ssubsec:learning_params}

For training the model, the learning rate was set to $5\times10^{-4}$, weight decay $9\times 10^{-4}$, batch size $64$, and maximum epoch number $150$ for post-processed models and $40$ for incrementally-trained models. The ADAM optimiser was used. A ten-fold cross-validation was used with a held-out test set for all data processed. Model weights were generated, stored and retrieved using the Python PyTorch library ~\cite{paszke2017automatic} with the TorchSparse extension~\cite{tang2022torchsparse}. 

The results using post-processed data were computed on a notebook Dell Precision 5700, \qty{64}{\giga\byte} RAM, Quadro RTX 5000 Mobile / Max-Q with \qty{16}{\giga\byte} video RAM. We aimed to achieve a training cycle every $\delta_t = 40s$ and have tuned the pipeline accordingly. This results in the collision map, the \ac{ograph} update and the online learning module running at the same rate. The empirically found duration of training cycles is a balance between generating sufficient new updates, allowing for the processing of the updates and accommodating overheads for the continuation of the training. 

\subsubsection{Online Graph Generation Parameters}
The local map of each node contains all the voxels that are contained in the cylinder defined up to $r_{max}=\qty{2}{\m}$. This is influenced by the robot's size and the patch size of previous work. The cylinder height is constrained by $z_{min} = \qty{-0.5}{\m}$ and $z_{max} = \qty{0.8}{\m}$ from the base link, with \qty{0.2}{\m} padding. 
\section{EXPERIMENTS AND RESULTS}
\label{sec:experiments}
The experiments in this section evaluate the online adaptability of the proposed TE method using data collected during operation and in situ on the robotic platform. Further, we aim to understand the performance and limitations of different adaptation or training strategies depending on the data available, e.g. post-processed vs online data. 

The experiments are structured in four parts. The initial set of experiments characterises how well the proposed method performs on different fine-tuning tasks on post-processed data. This provides insights into how well an existing model can be fine-tuned as well as a baseline for the online adaptation.
The second set of experiments demonstrates this approach in a real-world scenario, where the robot learns a new, untrained model using only the experience of an \qty{8}{\minute} data collection. It aims to validate the method by showing that it enables the robot to navigate autonomously in a densely vegetated environment.
The third set of experiments aims to train comparative methods and quantify their performance on the same data, avoiding any form of variation and stochastic elements, and compare three architecture variations and four different training setups.
The last set of experiments consists of a set of real-world experiments where differently trained models perform navigation experiments in varying environments, and their performance and limitations are assessed and compared. 

\subsection{Evaluation Metrics}
Throughout the evaluation, we use the Mathews Correlation Coefficient (MCC) score to evaluate and compare models. The MCC score measures how well the model predictions are correlated to the label data. It is immune to class swapping and is robust to imbalanced data sets. The MCC score ranges from $ -1$ to $1$, where 0 is a random model, and $1$ is a perfect positive correlation. It uses all four cases of the confusion matrix.  This makes it a preferred choice over the F1 score since the F1 score can overestimate a model's performance due to class imbalance, making it sensitive to class choice. This overestimation is particularly common for~\ac{TE}, where the positive class is typically the traversable class in the literature and is usually over-represented in training datasets. However, we also include the F1 score as a reference metric due to common practice in the literature.

\subsection{Data Sets and Data Set Groupings}
\label{subsec:data_set_groups}
\begin{figure}[ht]
    \centering
    \includegraphics[width=\columnwidth]{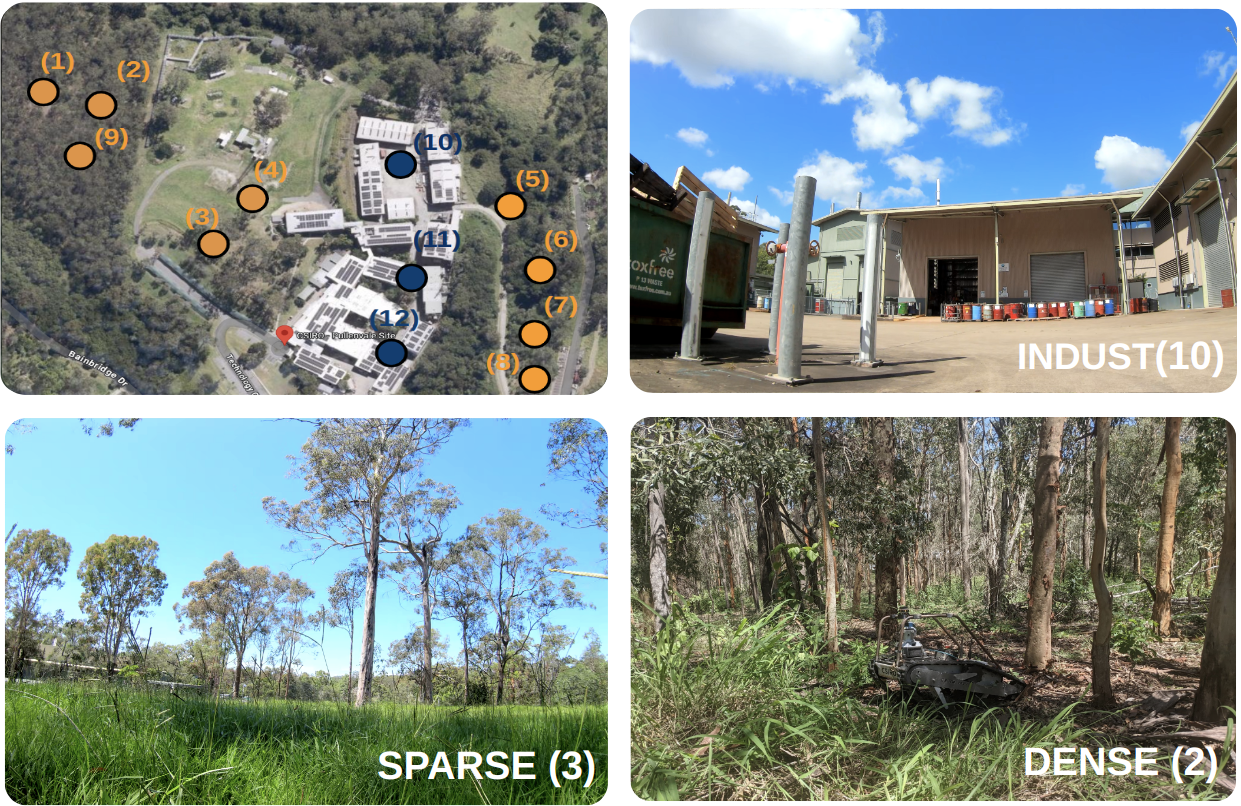}
    \caption{Top left: Shows the overview of the environment with the orange dots showing the locations of data sets from ForestTrav~\cite{ruetz2024foresttrav}. The blue dots are the novel data set introduced in this work from an industrial environment. An example is shown in the top right. The bottom left shows an example of the SPARSE environment, the bottom right scene from a densely vegetated forest.}
    \label{fig:data_set_environments}
\end{figure}

The data set used in this experiment was originally published by Ruetz et al.~\cite{ruetz2024foresttrav} and consists of 9 different scenes (numbers \# 1 to 9). These data sets capture a variation of open fields with grass, areas with open skies and small trees, and forests with no closed skies. The data set covers a variety of scenarios and different natural scenes. 
In this paper, we include three additional industrial data sets, numbered \# 10 to 12. These data sets are included to be used in a clearly different environment to the previously collected natural scenes and to validate adaptation to a different environment. These were collected using the \ac{dtr} robot at an industrial site of CSIRO in Pullenvale, Queensland, Australia. An overview of the data set and scenes can be found in Figure~\ref{fig:data_set_environments}, where the blue dots mark the locations of the newly collected data sets.

\subsubsection{Industrial Data Set Characterisation}
\label{subsec:industrial_data_set}
The industrial set contains external spaces found at the CSIRO QCAT site at Pullenvale, QLD, a mixed industrial area containing office buildings, workshops and large mechanical testing facilities, including open and closed spaces, illustrated in Figure~\ref{fig:data_set_environments}. The data sets numbered 1  and 12 are used for the training set, and 13 is the hold-out test data set. The first industrial data set, Scene \# 10, is an open industrial area surrounded by large sheds. The area has different obstacles, such as metal posts, guard rails and barrels. The second industrial scene (\# 11) contains a two-lane road with a tall shed on each side. A pedestrian walkway on the side inclines and declines in this scene, providing a narrow, challenging path. Additionally, there are a lot of guard rails, posts and fences that could prevent the robot from moving. The third scene consists of the same road but larger open areas and has obstacles similar to those in the first two scenes. This scene was used as a test set. For all these scenes, a lot of the traversable and non-traversable examples are similar. The traversable examples consist of different patches of flat surfaces with different inclinations. Different paths were taken to collect different inclinations. The non-traversable elements consist of posts, barrels, guard rails, containers, machinery and the large overall shed structures. In general, these elements are not as varied as what can be found in the scenes with vegetation.

The following Table~\ref{tab:industiral_data_set} provides an overview of each scene. The first column is the scene number, and the second column is the total number of voxels. The following four columns (2-6) provide the percentages of the hand-labelled and \ac{lfe} data for each traversability class. The seventh column provides the mean column density of a scene, called the column vegetation density (CVD)~\cite{ruetz2024foresttrav}. The density is the column-wise fraction, up to a pre-determined height, above and excluding the ground voxel containing measurements. The height is defined by the platform in this work and is set to \qty{1}{\m}. This indicates how much ``vegetation'' or other elements need to be considered and pushed through to navigate safely. The ForestTrav data set~\cite{ruetz2024foresttrav} averages densities of 0.4 - 0.57 versus 0.13 in the industrial scene. The last column is the scene dimensions in meters. 
 \begin{table*}
\caption{Overview of the industrial data set.}
\label{tab:industiral_data_set}
    \resizebox{\textwidth}{!}{
    \begin{tabular}{lccccccc}
    \toprule
    Scene & $N_{voxels}$ & $TR$ HL [\%] & $TR$ LfE  [\%]& $NTR$ HL  [\%]& $NTR$ LfE  [\%] &  Density & Dimensions [m] \\ \hline
    \# 10 & 187933 & 0.71 & 0.06 & 0.15 & 0.08 & 0.1    & $53.1 \times  69.72 \times 1.45$ \\
    \# 11  & 76497 & 0.56 & 0.08 & 0.32 & 0.04 & 0.1    & $46.67 \times 39.39 \times  1.85$ \\
    \# 12  &  69534 & 0.48 & 0.05 & 42 & 0.03 & 0.13  & $ 66.14 \times  40.63 \times  1.86 $ 
    \end{tabular}
}
\end{table*}

\subsubsection{Data Set Groupings}
Adaptation to unfamiliar and varying environments was investigated during the experiments and evaluation. Hence, the data sets were separated into three distinct groups, or subsets, by environment type: the industrial data set (INDUST), the sparse forest data set (SPARSE) and the dense forest data set (DENSE). The sparse forest contains data sets \# 3, 4, 5, 7 and 8, with number 8 being used as the test set. The sparse data set contains open fields with grass and bushes, as well as smaller trees with no overarching canopy. The increased sun exposure means there is a large mixture of dense vegetation near the ground and a few very large trees. Most of the vegetation near the ground is chlorophyll-rich. The dense forest data set contains scenes \# 1, 2, 6 as training data and \# 9 as test data. The dense forest contains significantly more underbrush and vegetation that does not contain chlorophyll, thin three stems and trunks. Additionally, there are overhanging branches, brambles and vines that make the environment more difficult. The visual difference can be seen in Figure~\ref{fig:data_set_environments}, where \# 3, 4 and 7 have little to no obstruction of the sky and contain large and small trees with open space. Comparatively, the dense forests \# 1 and 2 in Figure~\ref{fig:data_set_environments} are cluttered and have a higher tree density.
Later on, we refer to the ``complexity'' of the data set due to the environment. We consider the industrial environment to be the least complex and the dense, the most complex environment. The SPARSE data sets falls in between.

\subsubsection{Practical Considerations for LfE Data Collection}
\label{ch5:pratical_colmap}
As previously mentioned, the \ac{lfe} data collection was performed by an operator using an RC remote, recording the collision states. In practice, the operator ensured that the robot was driving (slowly) forward and collided only with the front of the robot with trees, bushes or other non-traversable elements. A collision was only recorded if the robot could not move physically further. Using a human operator induces a bias because the operator chooses how to collide with the elements in the environment, but this also helps to manage the risk of damaging the robot. The area was pre-determined, and the operator attempted to cover as much of it as possible in a single ``run'', revisiting places multiple times. During data collection, the operator collided with obstacles frequently and from many different directions. The more different, non-traversable collision states are captured, the higher the accuracy of the \ac{lfe} labelling approach. Note that mistakes during data collection are easy to make and that the overriding of the collision states can sometimes lead to erroneous self-labelling when using the heuristic method. 

\subsection{Model Adaptation on Post-processed Data}
\label{ch7:model_adpation_on_post_processed_data}
In the first set of experiments, a performance baseline was established for the proposed model and the different data sets due to the different labelling strategies. In the second sub-section, the adaptation or fine-tuning of post-processed data is explored. 

\subsubsection{Model Performance on Full Data Sets}
The initial set of experiments establishes baseline performance evaluation when using post-processed data from all the available data with different labelling strategies (LfE, LfE + HL). Table~\ref{tab:model_full_data_performance} shows the MCC scores and the F1 scores of the models. The first column specifies the labelling strategy, the second column is the MCC score, and the last column is the F1 score. The evaluation data set in all cases is scene \#9 with the \ac{lfe_hl} labelling strategy. The test set is the same as the one used in our prior work~\cite{ruetz2024foresttrav}. For consistency, all models are trained using the same number of epochs.

\begin{table}[h]
\centering
\caption{MCC Score for Models Trained on Post-Precessed Data}
\label{tab:model_full_data_performance}
\begin{tabular}{lcc}
\hline
Data Set &   MCC $ \mu \pm \sigma $ &  F1 $ \mu \pm \sigma$ \\
\hline    
LfE+HL    &     $0.71  \pm  0.03$    &       $0.85 \pm  0.017$   \\      
LfE        &    $0.69  \pm   0.022$     &       $0.85 \pm 0.012$   \\
\hline
\end{tabular}
\end{table}

Compared to our previous work~\cite{ruetz2024foresttrav}, there is an increase in \ac{mcc} score from 0.63 to 0.70~\ac{mcc} for the model trained on the LfE+HL data and the score obtained model trained on the \ac{lfe} data is only marginally lower.

\subsubsection{Fine-tuning on Post-Processed Data}
\label{ch7:finetuning_post_data}
In this experiment, we aim to evaluate the performance obtained when fine-tuning pre-trained models with data acquired in previously unseen environments. The first goal of this experiment is to confirm that models show different~\ac{TE} performance based on the different environment types due to the domain gap. The second goal is to explore whether fine-tuning models using data collected in a new target environment improves performance in the new environment. Given a fixed scaling and model architecture of a base model, we also aim to understand whether there is an upper bound for online adaptation, given a subset of hand-labelled and \ac{lfe} data for fine-tuning. In the context of online learning, this represents the case where a pre-trained base model is available, but the original training data is not. The base model can only be fine-tuned using novel data collected in the new environment.

\subsubsection{Base Models}
First, three base models (BM) are trained using data from one of the three groups defined previously only (industrial, sparse and dense forest), and each model is evaluated on test data of each of the three groups. The results are shown in Table~\ref{tab:offline_base}. The first column shows the base model names, and columns 2-6 show the MCC and F1 scores obtained with the trained model with respect to the test set of each of the groups. The MCC scores of models for which the training data and test set are from the same data set group are highlighted in italic text. The bold numbers indicate the best-performing models based on the MCC and F1 scores.

\begin{table}[h!]
\centering
\caption{Performance for Models Trained on Post-Processed Data Groups. Bold indicates best performance, and italic indicates matching training and test data groups.}
\label{tab:offline_base}
\begin{tabular}{@{}lllllll@{}}
\hline
\multicolumn{1}{l}{Base Models} & \multicolumn{2}{l}{ INDUST TS} & \multicolumn{2}{l}{SPARSE TS} & \multicolumn{2}{l}{DENSE TS} \\ \midrule
  & \multicolumn{1}{c}{{\color[HTML]{656565} \textit{MCC}}} & \multicolumn{1}{c}{{\color[HTML]{656565} \textit{F1}}} & \multicolumn{1}{c}{{\color[HTML]{656565} \textit{MCC}}} & \multicolumn{1}{c}{{\color[HTML]{656565} \textit{F1}}} & \multicolumn{1}{c}{{\color[HTML]{656565} \textit{MCC}}} & \multicolumn{1}{c}{{\color[HTML]{656565} \textit{F1}}} \\ \midrule
BM: INDUST &  \textbf{\emph{0.70}}  & \textbf{\emph{0.84}}  & 0.15  & 0.40 & 0.05 &0.12\\
BM: SPARSE & 0.63  &  0.79  &  \textbf{\emph{0.79}}  & \textbf{\emph{0.89}} & 0.56  & 0.78 \\
BM: DENSE & 0.69   &  0.84 & \textbf{0.79} &  \textbf{0.90} & \textbf{\emph{0.70}}  &  \textbf{\emph{0.84}} \\ \bottomrule
\end{tabular}%
\end{table}

An initial set of observations can be made on the base model's performance based on which subset of the data it is trained on.
\begin{itemize}
 \item The model trained on the DENSE data set (BM:DENSE) exhibits a marginally lower \ac{mcc} and similar F1 score when compared to the model trained on the full data set, with 0.69 vs 0.71 \ac{mcc} scores. This indicates that an accurate traversability predictor can be learned with a small amount of high-quality data that discriminates the traversable and non-traversable elements of the environment.
\item Secondly, models trained and evaluated on data from the same group show high \ac{mcc} scores. For example, a model trained on the SPARSE data set shows high scores on the SPARSE test set. These are the bold, italicised numbers.
\item We note that BM:DENSE shows comparable performance to models trained and evaluated on the data of the same group (italic numbers), indicating that the model generalised well over different environments. For example, when evaluated on the INDUST test set, the MCC is 0.70 for BM:INDUST and 0.69 for BM:DENSE. The generalisation of the model trained in the DENSE environment to the INDUST environments is surprising. This can be further seen qualitatively in Figure~\ref{fig:bm_genralisiation} and discussed in the subsection.
\item Lastly, base models trained on the industrial and sparse data sets are significantly less accurate when evaluated on the dense test set. This suggests that the DENSE data sets contained many unseen or complex elements not encountered in other environments. 
\end{itemize}

In general, we note that models trained on data from more ``complex environments'' seem to perform competitively in less complex environments as well. 

\subsubsection{Fine-tuning Base Models on Post-processed Data}
Next, each base model was fine-tuned using data from the three data set groups to train a fine-tuned model. This model was again tested with test sets from each group. This resulted in six permutations of training and fine-tuning pairs. We used the \ac{lfe} training data as this best resembles what the robot experiences in the field; it can be generated without time costly hand-labelling. The results are shown in Table~\ref{tab:tab_offline_ft}

The first column identifies the base model, e.g. BM:INDUST, which is the base model trained on the industrial data set. The second column defines the data set used to adapt (fine-tune) the base model. For ease of notation, we use ``AM'' (adapted model) to indicate which data set the model was fine-tuned with. The model was initially trained on the industrial set and fine-tuned on the dense data set, which is thus denoted by (BM:INDUST AM:DENSE). The other columns show the model's performance using MCC and F1 scores of the model for the test sets of the three groups. The numbers in parentheses are the scores of the base model, allowing us to see increases and decreases in the performance of the fine-tuned model versus the base model. Similar to the previous table, italics indicate the performance number for which the test set and the fine-tuning training data belong to the same group, and bold numbers indicate the highest performance for a given test set (per column).

Conceptually, the initial three rows in Table~\ref{tab:tab_offline_ft} are the cases where the complexity of the new environment (in the fine-tuning data set) is higher than experienced in the base training, e.g. from an industrial environment to a dense environment. For ease of terminology, we call this ``fine-tuned on more complex data''. 
Similarly, the last three rows for each architecture are the cases where the base model is trained on a higher-complexity data set and fine-tuned with lower-complexity data, e.g., the base model trained on the dense data set and fine-tuned on the industrial data set. For ease of terminology, we call these ``fine-tuned on less complex data''. 

The training was limited to 150 epochs to make sure the results were comparable to the online case, thereby setting an upper performance bound later for the online adaptation models. Table~\ref{tab:tab_offline_ft} shows the results.

\begin{table*}[h]
\caption{Performance For Fine-tuning Models }
\label{tab:tab_offline_ft}
\resizebox{\textwidth}{!}{%
\begin{tabular}{@{}llcccccc@{}}
\toprule
Base Model & Finetune Train Set & \multicolumn{2}{c}{INDUST TS} & \multicolumn{2}{c}{SPARSE TS} & \multicolumn{2}{c}{DENSE TS} \\ \midrule
 &  & {\color[HTML]{656565} \textit{MCC}} & {\color[HTML]{656565} \textit{F1}} & {\color[HTML]{656565} \textit{MCC}} & {\color[HTML]{656565} \textit{F1}} & {\color[HTML]{656565} \textit{MCC}} & {\color[HTML]{656565} \textit{F1}} \\ \midrule
                            & SPARSE& 0.40 (0.70) & 0.59 (0.84)  &\emph{ 0.32} (0.15) & \emph{0.57} (0.40) & 0.47 (0.05) & 0.73 (0.12) \\
 \multirow{-2}{*}{BM:INDUST} & DENSE & 0.74 (0.70) & 0.86 (0.84)  & 0.62 (0.15) & 0.80 (0.40) & \emph{0.54} (0.05) &\emph{ 0.74} (0.12) \\ \midrule
                             & DENSE & 0.71  (0.63) & 0.84 (0.79) & \textbf{0.80} 0.79) &\textbf{ 0.90} (0.89) & \emph{0.59 } (0.56) & \emph{0.79} (0.78) \\
\multirow{-2}{*}{BM: SPARSE} &INDUST &\emph{ 0.67}  (0.63) &\emph{ 0.82} (0.79) & 0.43 (0.79) & 0.69 (0.89) & 0.41  (0.56) & 0.66 (0.78) \\ \midrule
                            & INDUST &\textbf{\emph{ 0.81}} (0.69) &\textbf{ \emph{0.90}} (0.84) & 0.63 (0.79) & 0.78 (0.90) & 0.36 (0.70) & 0.55 (0.84) \\
\multirow{-2}{*}{BM: DENSE} & SPARSE & 0.49  (0.69) &0.67 (0. 4) & \emph{0.75} (0.79) & \emph{0.87} (0.90) &\textbf{ 0.65} (0.70) &\textbf{ 0.82} (0.84) \\ \bottomrule
\end{tabular}%
}
\end{table*}

For the fine-tuned models, we can make the following comments:
\begin{itemize}
 \item The ``fine-tuned on more complex data'' models are considerably more accurate than the base model. For example, the case (BM:INDUST, AM: DENSE) shows an increase in the accuracy of the model from \ac{mcc}=0.12 to 0.54. Generally, we note an overall increase in performance for all fine-tuned models that are ``fine-tuned on more complex data''. Further, for the dense environment, the maximum performance is \ac{mcc}=0.7 for the model initially trained on the dense and adapted with the sparse data set.   
 \item In the case where the base models are fine-tuned on less complex environments, the fine-tuning improves the model performance on the specific target environment. This increase can be marginal but comes at the cost of reducing the performance of the model in other environments, which is commonly known as ``catastrophic forgetting''. For example, pre-training the base model on sparse data and fine-tuning with the industrial data set increases the \ac{mcc} score from 0.63 to 0.67 on the industrial test set but decreases it from 0.79 to 0.43 on the sparse test data.
\end{itemize}

\begin{figure*}[h]
 \centering
 \includegraphics[width=\textwidth ]{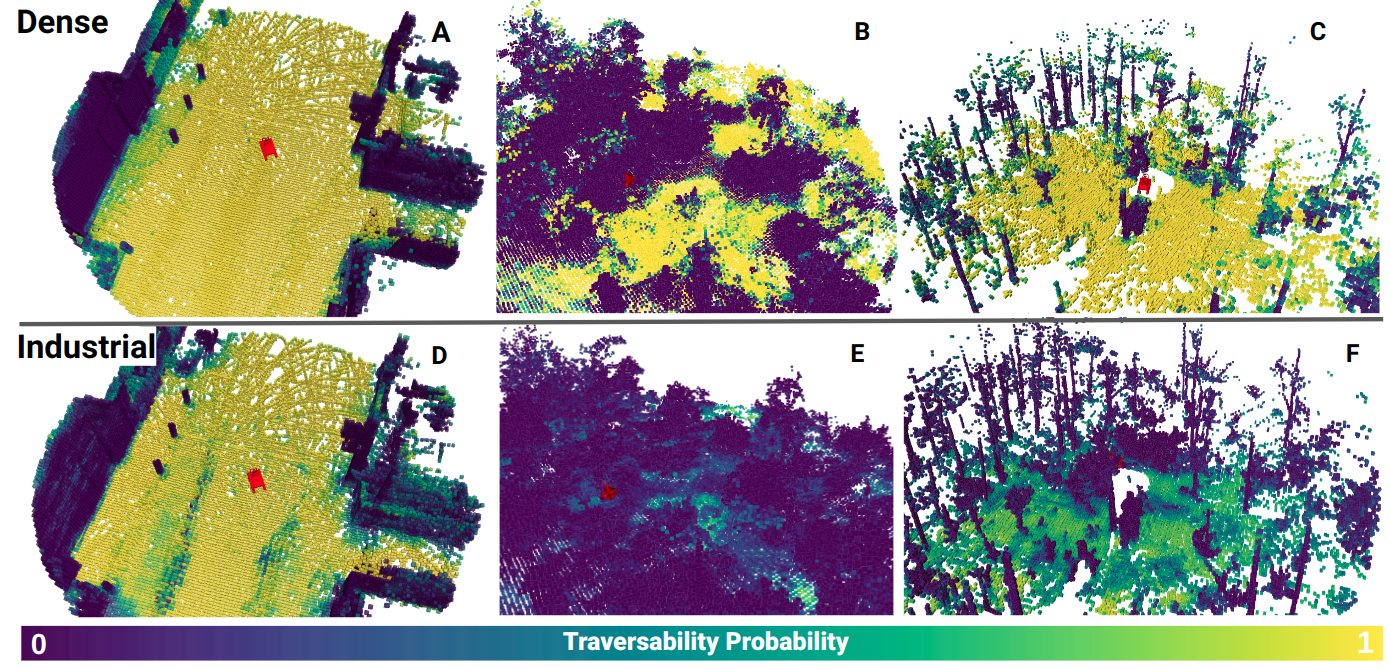}
 \caption[Generalisation of Models to Different Environments]{Comparison of two ensembles of models trained on either the dense data set in the top row (A, B, C) or the industrial data set in the bottom row (D, E, F). The left column shows each model's TE in an industrial environment, the middle and right columns show the TE of two densely vegetated environments with a variation of tree sizes and underbrush.}
 \label{fig:bm_genralisiation}
\end{figure*}

In the last paragraph, we provide some qualitative examples. Figure~\ref{fig:bm_genralisiation} shows a comparison between a base model trained in the industrial setting (bottom row) and one trained in the dense environment (top row) for different test environments (columns). Both use fused data sets, i.e.~\ac{lfe_hl} and are evaluated on the same environments using replayed online data from unseen environments. The left column is an industrial scene, the middle is from the sparse data set, and the right scenes are from dense forests. The sparse forest contains no canopy coverage, smaller and bushier trees, and open areas with significant chlorophyll-rich vegetation close to the ground.  In comparison, the dense forest contains tall trees of different tree diameters and significant chlorophyll-poor vegetation ( bramble, thin or young treed, etc) close to the ground.

The model trained on the dense data set generalises well to an industrial outdoor setting (A), assessing walls and smaller elements such as pylons correctly. This is a surprising and unexpected result. Learning-based systems trained in one environment commonly do not generalise to wholly different environments. For natural environments, other image-based methods have shown high sensitivity and degradation to small spatial location changes with similar environments~\cite{frey2023fast}.  Additionally, this model provides accurate~\ac{TE} for two different vegetated environments (B \& C).

The model trained on the industrial data set demonstrates accurate \ac{TE} for the industrial test environment (D) with some noise on the ground plane. The industrial model performs poorly when predicting traversability for vegetation close to the ground in (E). In (F), the model correctly assesses many of the large elements (trees) but struggles with the smaller elements. However, there is a clearly visible colour gradient (dark blue to light green) on some of the smaller elements, showing discrimination of the \ac{TE} of the vegetation near the ground, clearly identifying tree trunks as non-traversable but struggling with bushes and smaller forms of vegetation. The model has learned a representation of the industrial setting, but it does not generalise to the densely vegetated environments. 
Comparatively, the model trained on the dense data set exhibits high performance over all three scenes, even the industrial one.

\subsection{Real-World Demonstration of Online Learning and Navigation in Forest}
\label{subsec:online_odap}

This experiment demonstrates that the proposed method can train a model on the robot itself, guided by a human operator in situ. The model was trained onboard the robot on an NVIDIA Jetson Orin NX (25W). A newly trained model was qualitatively demonstrated by point-to-point navigation and compared against the test set \#9.  The model was randomly initialised (untrained, new model), and it was continuously fine-tuned with incrementally acquired \ac{lfe} data. This corresponds to the case (BM 0, FT 1) described in sub-section~\ref{subsec:quant_online_adapt}. The training data collection was completed in less than 8 minutes and resulted in an \ac{mcc} score of 0.63, evaluated on the test scene \#9.

\begin{figure}[h]
 \centering
 \includegraphics[width= .9\linewidth]{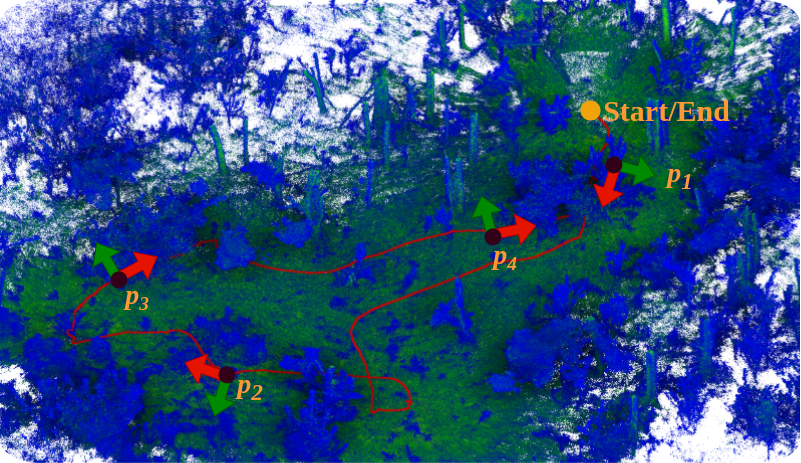}
 \caption[Overview of Robot Navigation of the Online Learned Model]{Overview of the navigation in a forest environment, with poses $\mathbf{p}_1$ - $\mathbf{p}_4$. The navigation was completed with the O-line trained model in situ.}
 \label{fig7:odap_nav_overview}
\end{figure}

The model was deployed on the robot and successfully navigated a closed-loop point-to-point trajectory between three waypoints, $w_1$, $w_2$, and the initial start location. A visualisation of the path can be seen in~\figref{fig7:odap_nav_overview}. The poses $p1$-$p4$ are visualisations of the trajectory, where the left image is the RGB FPV view of the robot, the middle image is the 3D TE estimation, and the right image is the costmap. Details on the costmap are provided in Section~\ref{subsec:cotmap_generation}.

\begin{figure*}[ht!]
 \centering
 \begin{subfigure}{0.95\linewidth}
 \centering
 \includegraphics[width=\linewidth]{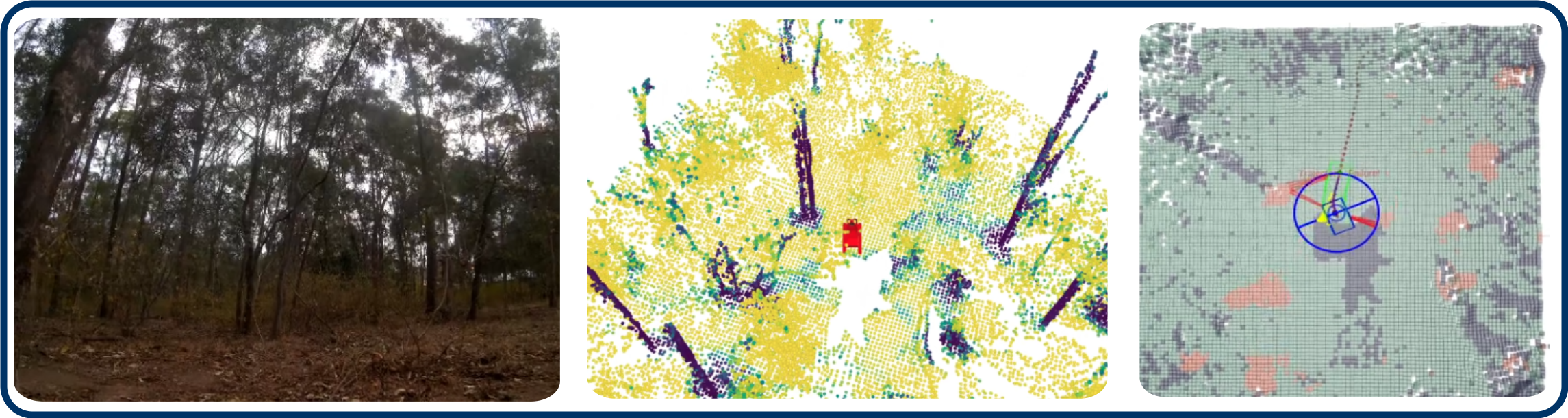}
 \caption{Robot front camera (left), 3D TE estimation (middle), costmap (right) for pose $\mathbf{p}_1$}
 \label{fig7:online_p1}
 \end{subfigure}
 \hfill
 \begin{subfigure}{0.95\linewidth}
 \centering
 \includegraphics[width=\linewidth]{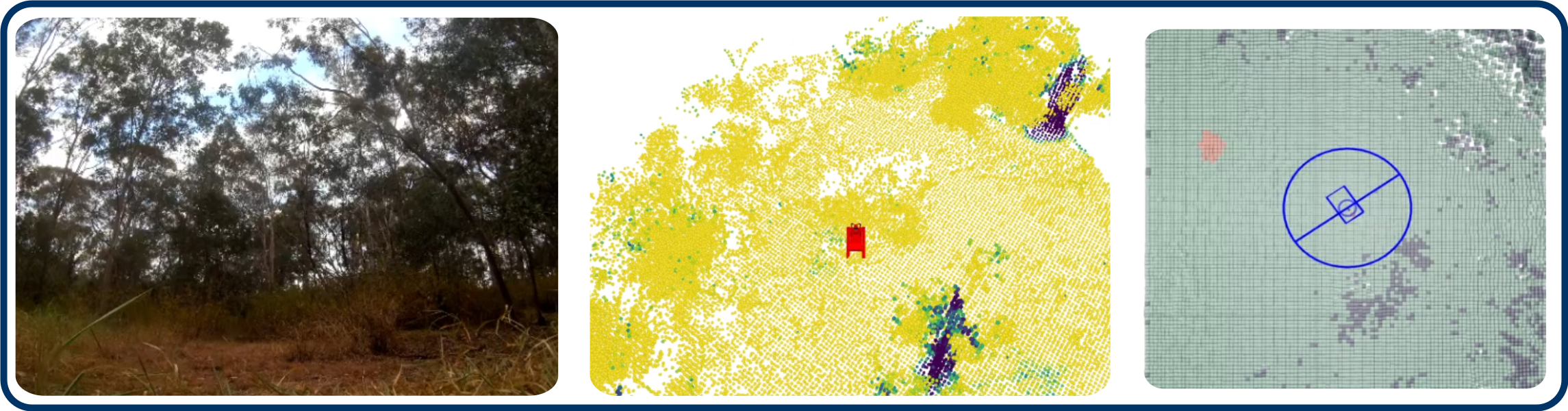}
 \caption{Robot front camera (left), 3D TE estimation (middle), costmap (right) for pose $\mathbf{p}_2$}
 \label{fig7:online_p2}
 \end{subfigure}
 \begin{subfigure}{0.95\linewidth}
 \centering
 \includegraphics[width=\linewidth]{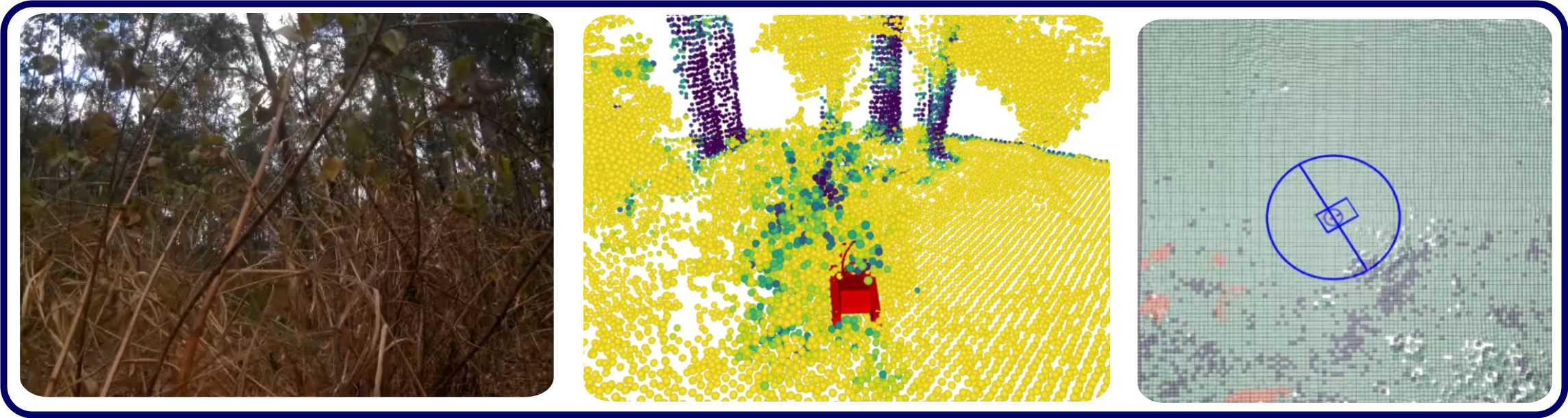}
 \caption{Robot front camera (left), 3D TE estimation (middle), costmap (right) for pose $\mathbf{p}_3$}
 \label{fig7:online_p3}
 \end{subfigure}
 \begin{subfigure}{0.95\linewidth}
 \centering
 \includegraphics[width=\linewidth]{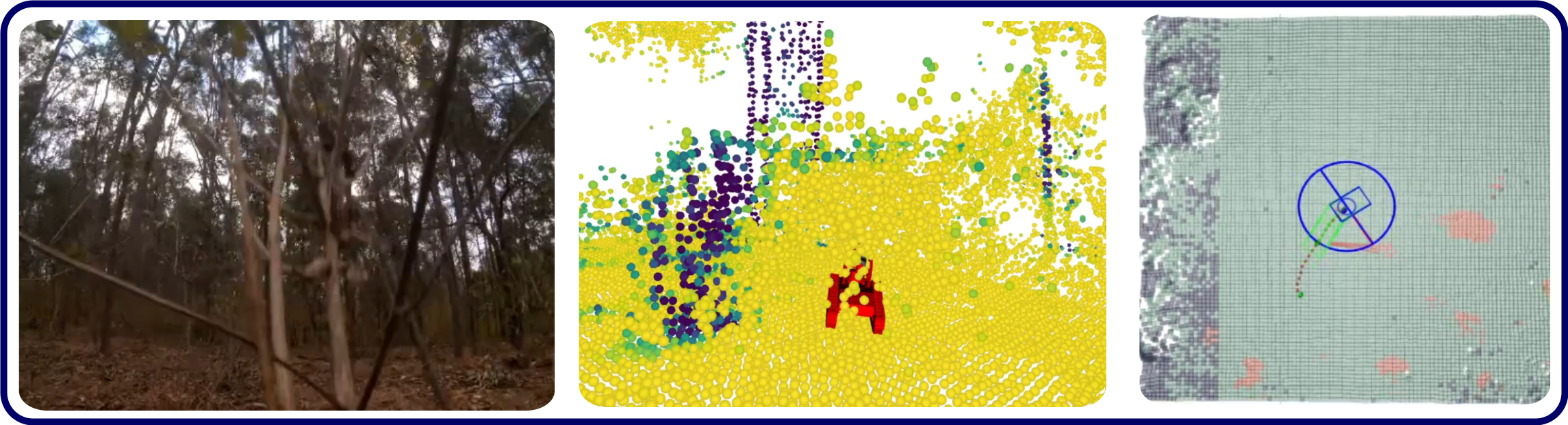}
 \caption{Robot front camera (left), 3D TE estimation (middle), costmap (right) for pose $\mathbf{p}_4$}
 \label{fig7:online_p4}
 \end{subfigure}
 \caption[Scenes from Autonomous Robot Navigation using Online TE Approach]{Visualisation scenes from a successful online point-to-point navigation using the online learnt TE model in target environments, $\mathbf{p}_1$ - $\mathbf{p}_4$ are poses along the trajectory.}
 \label{fig7:odap_online_nav}
\end{figure*}

Subfigure~\ref{fig7:online_p1} shows the robot at the beginning of the trajectory briefly after the initialisation. The robot correctly assesses the trees and smaller vegetation elements, allowing it to push through the initial stand of bushes. There are some areas with uncertain (green) TE estimates, likely due to few observations; see comments for $\mathbf{p}_4$.

For $\mathbf{p}_2$, the robot encountered a large patch of grass that reached the camera. The proposed online learned TE correctly assessed it as traversable and pushed through it whilst avoiding the unobserved areas.

The pose $\mathbf{p}_3$ shows how the robot pushed through a thick bramble bush and proceeded to continue. There were some possibly non-traversable elements higher up in the bush. These elements were not projected onto the 2D costmap since they were above the robot's height threshold, and the robot would not interact with them (see Subsection~\ref{subsec:cotmap_generation}). In the background, it can be clearly observed that the trees were correctly assessed as non-traversable.

 At pose $\mathbf{p}_4$, the robot navigated through the bramble/small trees to reach the initial starting location, shown in Subfigure~\ref{fig7:online_p4}. The robot avoided the thicker tree to the right and reached the starting location. Compared to $\mathbf{p}_1$, the local costmap was less uncertain for the 3D TE estimate. The area has been observed multiple times and at multiple angles, and this allowed for more accurate traversability estimation than initially. In the visualisation of the costmap, one can observe a band on the left of traversable and unobserved elements that have been observed at the start of the experiment. This is an example of the de-allocation of the local probabilistic map since the regions were out of bounds, which is why the graph is needed.

In summary, the experiment demonstrated that a new model can be trained in situ, reaching accurate performance (\ac{mcc}$=0.63$), and is sufficient for safe point-to-point navigation in densely vegetated environments. Further, the model performance is comparable to the performance of the off-line method reported in our latest publication~\cite{ruetz2024foresttrav}.

\subsection{Quantification of Online Adaptation}
\label{subsec:quant_online_adapt}

The primary question this experiment addresses is whether the model can be adapted/trained online with only data gathered in a relevant environment. Secondly, it identifies the best strategy to do so. For this purpose, a data set was collected with human-assisted collisions. The online data set was replayed and stored in \qty{40}{\s} increments using the \ac{ograph} to generate snapshots of the online experience, which can be used to train. The intermediate data storing step was chosen to ensure data consistency.

\begin{figure*}[ht!]
 \centering
 \includegraphics[width=\textwidth ]{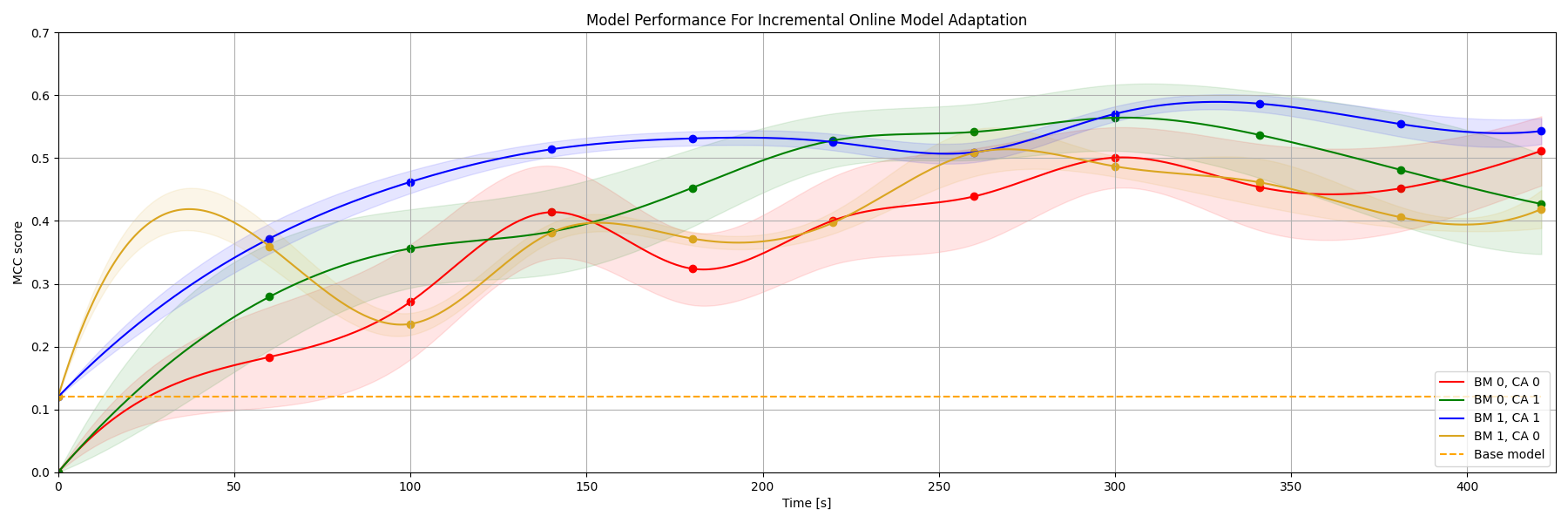}
 \caption{Comparison of the performance of different models over the incremental online adaptation for the four different cases. The $x$-axis shows the time, and the $y$-axis the \ac{mcc} score. The transparent regions of the graph correspond to the upper and lower bounds for one standard deviation.}
 \label{fig:aol_incr_perf}
\end{figure*}

We examine four training strategies. They are defined by the two boolean flags BM (use of a base model) and CA (continuous adaptation). The four variations of these strategies can be seen in \figref{fig:aol_incr_perf}. If the BM flag is set to true (BM 1), the method uses an initial base model trained on the industrial data set. Otherwise (BM 0), the model is randomly initialised. The industrial base model was chosen since the environment is the most different from the dense forest. For models trained without any prior knowledge, pre-determined scaling values were used to ensure non-catastrophic scaling during the run. The CA flag denotes if, for each training cycle, \ac{ograph} is updated, the model trained on the previous iteration is used for the next training cycle. If CA is true (CA 1), the model is continuously updated as new information comes in. If CA is false (CA 0), the model is trained from the base model at each ``training cycle'', as defined above.

This results in four distinct cases in \figref{fig:aol_incr_perf}:
\begin{enumerate}
 \item BM 0, CA 0 (red) -- randomly-initialised base model retrained at each training cycle,
 \item BM 0, CA 1 (green) -- randomly-initialised base model continuously adapted,
 \item BM 1, CA 1 (blue) -- pre-trained base model continuously adapted, and
 \item BM 1, CA 0 (gold) -- pre-trained base model retrained at each training cycle.
 \end{enumerate}
The dashed yellow line shows the base-model performance (no retraining).

The data comes from a newly collected data set with the goal of maximising collisions (the minority class) within a new area. The test data set remains scene \# 9 from the previously established data set, allowing for a comparison to all other experiments.

Looking at \figref{fig:aol_incr_perf}, the performance of the individual models reaches a ceiling between 0.55 and 0.65 \ac{mcc} score at their peak, the mean of the models just below 0.6~\ac{mcc} score. All models learn and improve substantially throughout online adaptation. There is a clear suggestion that a model can be either adapted or trained online with only data collected through experience and that this can be achieved in real time on a robotic platform in situ.
The variant (BM 1, CA 1) (blue graph) is generally the highest performing and most consistent (lowest variance) of all three models. The lower variance cannot solely be attributed to the learning rate being lower for continual adaptation, and we see higher variation for the green plots. The models with no continual learning (CA 0), red and yellow, show an oscillation effect, where the model's performance sometimes decreases over iterations. Further, higher variation of the method for models with randomly-initialised base models (BM 0), red and green, can be observed. 

\subsection{Navigation in Different Environments}
\label{subse:nav_in_different_env}
In this investigation, we comparatively assess navigation performance across multiple methodological approaches in two distinct forest environments. The comparative analysis includes NavStack~\cite{HudTal21}, a traditional (rigid world assumption) geometric lidar-based occupancy method, baseline ForestTrav trained on post-processed datasets~\cite{ruetz2024foresttrav}, ForestTrav DENSE and ForestTrav INDUST (trained on dense and industrial datasets, respectively), and the proposed Adaptive Online Learning (AOL) method. The experimental evaluation was conducted post-AOL generation, utilising a platform equipped with a different instance of a Velodyne VLP-16 lidar from previous dataset acquisition platforms, known to have shifts in intensity characteristics.

For two locations, a pre-defined starting and end point were given, and each method attempted to reach its goal. If a model got stuck for the first time, the operator would intervene by moving the robot forward and then continuing the experiment. For the first experiment, the robot was sent \qty{20}{\m} ahead but had to pass through a wall of small plants, navigate through an open space with sparse vegetation and finally navigate around a variation of high grass and small trees. 
For the second experiment, the robot had to navigate to a goal pose located \qty{40}{\m} from the starting position. A combination of cluttered, small trees with a variation of underbrush as well as fallen tree trunks had to be avoided.

In the first experiment, ForestTrav AOL and the original ForestTrav successfully reached the end goal without any intervention. ForestTrav initially struggled with bushes but managed to navigate around them, while ForestTrav AOL took a more direct route through the foliage. ForestTrav DENSE failed to bypass the bushes and required operator intervention to continue. NavStack also failed at the start, and after intervention, diverged from the goal due to avoiding small stems, ultimately failing.

In the second experiment, ForestTrav and ForestTrav AOL completed the course again without any human intervention. ForestTrav DENSE and NavStack made some initial progress through cluttered trees but became stuck due to thin stems and clutter, with intervention proving ineffective. ForestTrav INDUST failed entirely, unable to assess the environment.

ForestTrav and ForestTrav AOL were the most effective methods, differing primarily in their interpretation of traversable terrain. ForestTrav struggled with vegetation that was absent in its training data but prevalent in the target environment. ForestTrav AOL, trained in a similar environment, handled these obstacles better but was less robust overall due to limited training data and the lack of an ensemble approach. The poor performance of ForestTrav DENSE is attributed to the insufficient representation of thin, bushy vegetation in its training set.

ForestTrav INDUST consistently failed, likely due to changes in the intensity distribution characteristics of the Velodyne VLP-16 sensor. This was confirmed in follow-up tests, but the costmaps remained inadequate for navigation.

\begin{figure*}
     \centering
    \includegraphics[width=\textwidth ]{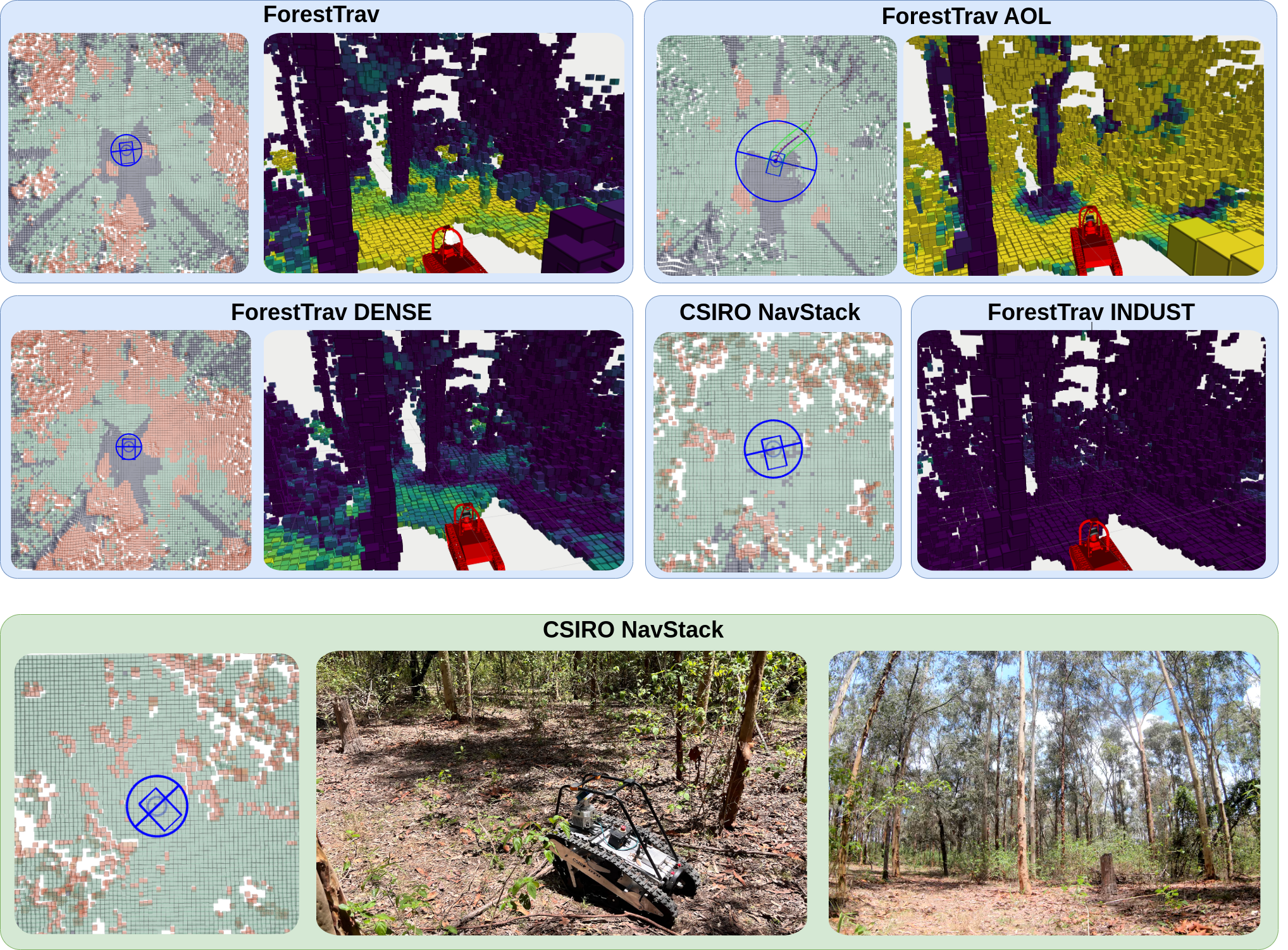}
    \caption{Qualitative examples of the methods in the target environment. Blue blocks are for all methods at location 1 and show the starting area of the experiment, allowing us to compare the methods. The green block shows the costmap, FPV and external view of the robot for the location where NavStack failed to navigate. }
    \label{fig:nav_experiment}
\end{figure*}

\begin{table}[t]
\caption{Comparison performance of different methods for different navigation environments. }
\label{tab:nav_comp_exp}
\resizebox{\columnwidth}{!}{%
\begin{tabular}{@{}l|ccc|ccc@{}}
\toprule
Method & \multicolumn{3}{c|}{Location 1}  & \multicolumn{3}{c}{Location 2} \\ \midrule
 & \multicolumn{1}{r}{{\color[HTML]{656565} \textit{Success}}} & \multicolumn{1}{r}{{\color[HTML]{656565} \textit{Time [s]}}} & \multicolumn{1}{r|}{{\color[HTML]{656565} \textit{Distance [m]}}} & \multicolumn{1}{r}{{\color[HTML]{343434} \textit{Success}}} & \multicolumn{1}{r}{{\color[HTML]{343434} \textit{Time [s]}}} & \multicolumn{1}{r|}{{\color[HTML]{343434} \textit{Distance [m]}}} \\ \midrule
CSIRO NavStack & No & 240 & 30.7 & No & 170 & 13.9  \\
ForestTrav & Yes & 270 & 32.1 & Yes & 350 & 52.1  \\
ForestTrav AOL & Yes & 160 & 27.1 & Yes & 300 & 52.1\\
ForestTrav DENSE & No & 190 & 24.5 & No & 220 & 20.29  \\
ForestTrav INDUST & No & - & 0.0 & No & -  & 0.0 \\
\end{tabular}%
}
\end{table}

\section{DISCUSSION}
\label{sec:discussion}
The online adaptation demonstrated on the real-world data set showed that thanks to our proposed method, it is feasible to train and adapt a model online and deploy it in a vegetated environment for safe navigation. This is a significant finding since, currently, there are no comparable online~\ac{TE} adaptation methods for robotic navigation in vegetated environments using probabilistic voxel representations. Further, we demonstrated the method can achieve high performance, comparable to the results obtained in state of the art using offline data~\cite{ruetz2024foresttrav}. 

Further, the presented results on online adaptation provided insights and recommendations for practical applications for deploying robots using deep learning systems and probabilistic voxel representations. The performance-determining factors are the quality and quantity of available data. Depending on time, cost and performance constraints, there are different recommendations. For a long-term deployment in a given area, the traditional hand-labelling approach provides the best performance at a high upfront cost and with the most generalisation. 

However, the results from the experiments also suggest that highly accurate models can be trained with data solely collected from the robot's experience. Multiple data collection runs may be required, and for the presented system and methods, this can be achieved within hours. However, as in the case with the ForestTrav DENSE case, the model may be only valid for a short time and may not generalise as well to changing sensor or environmental changes.

The online adaptation on the robot could be reduced to \qty{\sim 8}{\minute} for our system running on a \qty{25}{\watt} GPU, with a previous model at the reduction of MCC score from 0.74 to 0.63 for our in-situ online adaptation experiment, see Subsection~\ref{subsec:online_odap}. Comparing the adaptive online method score with results from Table~\ref{tab:tab_offline_ft}, the online approach is not as accurate as the \ac{lfe_hl} methods. However, the online experiment~\ref{subsec:online_odap} resulted in a model as accurate, both MCC of 0.63, as the models trained only~\ac{lfe} the available time, specific task and accuracy requirements will dictate what method is appropriate.

The main difference between the pure \ac{lfe} data set (post-processed data) and the online adaptation data set is the information quality of the voxel map. Again, understanding the completeness of the environment and how well the voxel-wise distributions have converged is desirable, but it is a current limitation. Using offline processing allows us to store intermediate steps, and introspection is currently not possible when computing online. Further, the proposed models require, in comparison to image methods, little high-quality data to perform well (see Table~\ref{tab:offline_base}). This indicates that the quality and variety of the data are more significant than the necessary quantity of data. The experiment using the online collected data showed that high-performance models could be trained. However, there must be sufficient and diverse data.

A key benefit of this method is its generalisation over starkly different environments. The results indicate that the more complex environments share many baseline similarities with structured environments, such as the industrial set, but include many more complex elements (e.g. grass, bushes). However, some baseline static constraints will still apply, e.g. an agent cannot drive up to steep slopes or cannot overcome large elements. Other online, adaptive vision-based systems have reported significant performance degradation within small location changes of the ``same'' environment, noting this is common and expected for vision modalities~\cite{frey2023fast}.  

Whilst our method generalises over different environments, in our online navigation experiment, we see a qualitative performance degradation of the model trained on the ForestTrav data set. Firstly, the test environment has experienced significant growth since the original data set~\cite{ruetz2024foresttrav} was collected, changing the environment substantially. The presented online adaptive model, ForestTrav AOL, generated its training data in the changed environment, highlighting the need for adaptive methods. Secondly, a different sensor pack was used during testing, changing the characteristics of some of the sensor modalities, leading to significant degradation of the model trained on the industrial setting in the forest environment.



\section{CONCLUSION}
This work introduced a novel online TE adaptation method that allows a robot to adapt to its environment in situ. In human-robot collaboration, the robot experiences the environment and can build high-quality training data on the fly using a graph-based representation \ac{ograph}. We explored the lidar-only TE estimation model for online adaptation to a novel environment. We demonstrated the feasibility of adapting a model online. The proposed methods and \ac{ograph} allow the model's training in situ whilst aided by a human operator. We demonstrated that the adapted model allowed for point-to-point navigation in challenging environments. The online-trained model with an \ac{mcc} score of 0.63 allows the robot to navigate complex environments and shows comparable performance to the model trained on offline data in our previous publication~\cite{ruetz2024foresttrav}.

We compare the benefits of online adaptation versus more classical, offline approaches using hand-labelled data or self-supervised data, allowing potential future users to make an informed decision about which method to choose.  ForestTrav, using the combination of hand and self-supervised data, has been shown to generalise well to novel environments and environmental changes.  However, the presented online adaptive method has significant practical use benefits given time constraints. Generating training data and a trained model in less than \qty{8}{\minute} in situ is much faster than post-processing and hand-labelling data. Lastly, since there is limited literature on online learning with 3D probabilistic voxel on robotic platforms in situ, this work provides valuable insights and highlights further challenges and open research areas.

 Current limitations include ensuring that the training data is unique and expressive. The longer an adaptation lasts, the more similar data will likely be accumulated. The increase in duplicate data reduces the training signal of corner cases due to the low number of samples. Hence, understanding what is poorly observed vs out-of-distribution or novel data would be an interesting avenue of future work. In an ideal case, this could be presented to the operator in real-time to collect relevant information and allow the model to adapt faster. Furthermore, parallel training of models has yet to be explored. Simultaneously, adapting multiple models would be desirable to enable the use of an ensemble. Lastly, a method to evaluate the model online during deployment, either fully trained or adapted, would be desirable.

\bibliographystyle{IEEEtran}      
\bibliography{References}       

\begin{thebibliography}{10}
\providecommand{\url}[1]{#1}
\csname url@samestyle\endcsname
\providecommand{\newblock}{\relax}
\providecommand{\bibinfo}[2]{#2}
\providecommand{\BIBentrySTDinterwordspacing}{\spaceskip=0pt\relax}
\providecommand{\BIBentryALTinterwordstretchfactor}{4}
\providecommand{\BIBentryALTinterwordspacing}{\spaceskip=\fontdimen2\font plus
\BIBentryALTinterwordstretchfactor\fontdimen3\font minus \fontdimen4\font\relax}
\providecommand{\BIBforeignlanguage}[2]{{%
\expandafter\ifx\csname l@#1\endcsname\relax
\typeout{** WARNING: IEEEtran.bst: No hyphenation pattern has been}%
\typeout{** loaded for the language `#1'. Using the pattern for}%
\typeout{** the default language instead.}%
\else
\language=\csname l@#1\endcsname
\fi
#2}}
\providecommand{\BIBdecl}{\relax}
\BIBdecl

\bibitem{frey2023fast}
J.~Frey, M.~Mattamala, N.~Chebrolu, C.~Cadena, M.~Fallon, and M.~Hutter, ``Fast traversability estimation for wild visual navigation,'' \emph{arXiv preprint arXiv:2305.08510}, 2023.

\bibitem{guastella2021learning}
D.~C. Guastella and G.~Muscato, ``Learning-based methods of perception and navigation for ground vehicles in unstructured environments: A review,'' \emph{Sensors}, vol.~21, no.~1, p.~73, 2021.

\bibitem{bradley2015scene}
D.~M. Bradley, J.~K. Chang, D.~Silver \emph{et~al.}, ``Scene understanding for a high-mobility walking robot,'' in \emph{IROS}, 2015.

\bibitem{bae2023self}
J.~Bae, J.~Seo, T.~Kim, H.-G. Jeon, K.~Kwak, and I.~Shim, ``Self-supervised 3d traversability estimation with proxy bank guidance,'' \emph{IEEE Access}, vol.~11, pp. 51\,490--51\,501, 2023.

\bibitem{wellhausen2019where}
L.~Wellhausen, A.~Dosovitskiy, R.~Ranftl \emph{et~al.}, ``Where should {I} walk? predicting terrain properties from images via {Self-Supervised} learning,'' \emph{IEEE RAL}, vol.~4, no.~2, 2019.

\bibitem{kahn2020badgr}
G.~Kahn, P.~Abbeel, and S.~Levine, ``{BADGR}: An autonomous self-supervised learning-based navigation system,'' \emph{IEEE RAL}, vol.~6, no.~2, 2021.

\bibitem{li2023seeing}
A.~Li, C.~Yang, J.~Frey, J.~Lee, C.~Cadena, and M.~Hutter, ``Seeing through the grass: Semantic pointcloud filter for support surface learning,'' \emph{arXiv preprint arXiv:2305.07995}, 2023.

\bibitem{frey2024roadrunner}
J.~Frey, S.~Khattak, M.~Patel, D.~Atha, J.~Nubert, C.~Padgett, M.~Hutter, and P.~Spieler, ``Roadrunner--learning traversability estimation for autonomous off-road driving,'' \emph{arXiv preprint arXiv:2402.19341}, 2024.

\bibitem{yoon2024adaptive}
H.-S. Yoon, J.-H. Hwang, C.~Kim, E.~I. Son, S.-W. Yoo, and S.-W. Seo, ``Adaptive robot traversability estimation based on self-supervised online continual learning in unstructured environments,'' \emph{IEEE Robotics and Automation Letters}, 2024.

\bibitem{papadakis2013terrain}
P.~Papadakis, ``Terrain traversability analysis methods for unmanned ground vehicles: A survey,'' \emph{Engineering Applications of Artificial Intelligence}, vol.~26, no.~4, 2013.

\bibitem{HudTal21}
N.~Hudson, F.~Talbot, M.~Cox, and other, ``Heterogeneous ground and air platforms, homogeneous sensing: {T}eam {CSIRO} {D}ata61's approach to the {DARPA} {S}ubterranean {C}hallenge,'' \emph{Field Robotics J.}, 2022.

\bibitem{tranzatto2022cerberus}
M.~Tranzatto, F.~Mascarich, L.~Bernreiter, C.~Godinho, M.~Camurri, S.~Khattak, T.~Dang, V.~Reijgwart, J.~Loeje, D.~Wisth \emph{et~al.}, ``Cerberus: Autonomous legged and aerial robotic exploration in the tunnel and urban circuits of the darpa subterranean challenge,'' \emph{arXiv preprint arXiv:2201.07067}, p.~3, 2022.

\bibitem{wellington2004online}
C.~Wellington and A.~Stentz, ``Online adaptive rough-terrain navigation vegetation,'' in \emph{{IEEE} International Conference on Robotics and Automation, 2004. Proceedings. {ICRA} '04. 2004}, vol.~1, Apr. 2004, pp. 96--101.

\bibitem{chen2023rspmp}
D.~Chen, M.~Zhuang, X.~Zhong, W.~Wu, and Q.~Liu, ``Rspmp: Real-time semantic perception and motion planning for autonomous navigation of unmanned ground vehicle in off-road environments,'' \emph{Applied Intelligence}, vol.~53, no.~5, pp. 4979--4995, 2023.

\bibitem{quigley2009ros}
M.~Quigley, K.~Conley, B.~Gerkey, J.~Faust, T.~Foote, J.~Leibs, R.~Wheeler, A.~Y. Ng \emph{et~al.}, ``Ros: an open-source robot operating system,'' in \emph{ICRA workshop on open source software}, vol.~3, no. 3.2.\hskip 1em plus 0.5em minus 0.4em\relax Kobe, Japan, 2009, p.~5.

\bibitem{fan2021step}
D.~D. Fan, K.~Otsu, Y.~Kubo \emph{et~al.}, ``{STEP}: Stochastic traversability evaluation and planning for risk-aware off-road navigation,'' \emph{arXiv preprint arXiv:2103.02828}, 2021.

\bibitem{ahtiainen2017normal}
J.~Ahtiainen, T.~Stoyanov, and J.~Saarinen, ``Normal distributions transform traversability maps: {LIDAR}-only approach for traversability mapping in outdoor environments,'' \emph{Journal of Field Robotics}, vol.~34, no.~3, 2017.

\bibitem{ruetz2022FTM}
F.~Ruetz, P.~Borges, N.~Suenderhauf, E.~Hern{\'a}ndez, and T.~Peynot, ``Forest traversability mapping ({FTM}): Traversability estimation using {3D} voxel-based normal distributed transform to enable forest navigation,'' in \emph{IROS}, 2022.

\bibitem{ruetz2024foresttrav}
F.~Ruetz, N.~Lawrance, E.~Hern{\'a}ndez, P.~Borges, and T.~Peynot, ``Forest{T}rav: 3{D} lidar-only forest traversability estimation for autonomous ground vehicles,'' \emph{IEEE Access}, 2024.

\bibitem{kim2006traversability}
D.~Kim, J.~Sun, S.~M. Oh, J.~M. Rehg, and A.~F. Bobick, ``Traversability classification using unsupervised on-line visual learning for outdoor robot navigation,'' in \emph{Proceedings 2006 IEEE International Conference on Robotics and Automation, 2006. ICRA 2006.}\hskip 1em plus 0.5em minus 0.4em\relax IEEE, 2006, pp. 518--525.

\bibitem{hadsell2009learning}
R.~Hadsell, P.~Sermanet, J.~Ben, A.~Erkan, M.~Scoffier, K.~Kavukcuoglu, U.~Muller, and Y.~LeCun, ``Learning long-range vision for autonomous off-road driving,'' \emph{Journal of Field Robotics}, vol.~26, no.~2, pp. 120--144, 2009.

\bibitem{hornung2013octomap}
A.~Hornung, K.~M. Wurm, M.~Bennewitz, C.~Stachniss, and W.~Burgard, ``{OctoMap}: An efficient probabilistic 3d mapping framework based on octrees,'' \emph{Autonomous robots}, vol.~34, pp. 189--206, 2013.

\bibitem{saarinen2013normal}
J.~Saarinen, H.~Andreasson, T.~Stoyanov, J.~Ala-Luhtala, and A.~J. Lilienthal, ``Normal distributions transform occupancy maps: Application to large-scale online 3d mapping,'' in \emph{2013 IEEE international conference on robotics and automation}.\hskip 1em plus 0.5em minus 0.4em\relax IEEE, 2013, pp. 2233--2238.

\bibitem{ramezani2022wildcat}
M.~Ramezani, K.~Khosoussi, G.~Catt \emph{et~al.}, ``Wildcat: Online continuous-time {3D} lidar-inertial {SLAM},'' \emph{arXiv preprint arXiv:2205.12595}, 2022.

\bibitem{uzawa2022end}
Y.~Uzawa, S.~Matsuzaki, H.~Masuzawa, and J.~Miura, ``End-to-end path estimation and automatic dataset generation for robot navigation in plant-rich environments,'' in \emph{International Conference on Intelligent Autonomous Systems}.\hskip 1em plus 0.5em minus 0.4em\relax Springer, 2022, pp. 272--284.

\bibitem{tang2022torchsparse}
H.~Tang, Z.~Liu, X.~Li, Y.~Lin, and S.~Han, ``{TorchSparse: Efficient Point Cloud Inference Engine},'' in \emph{Conference on Machine Learning and Systems (MLSys)}, 2022.

\bibitem{good1952rational}
I.~J. Good, ``Rational decisions,'' \emph{Journal of the Royal Statistical Society: Series B (Methodological)}, vol.~14, no.~1, pp. 107--114, 1952.

\bibitem{dolgov08}
D.~Dolgov, S.~Thrun, M.~Montemerlo, and J.~Diebel, ``Practical search techniques in path planning for autonomous driving,'' in \emph{Proceedings of the First International Symposium on Search Techniques in Artificial Intelligence and Robotics (STAIR-08)}.\hskip 1em plus 0.5em minus 0.4em\relax Chicago, USA: AAAI, June 2008.

\bibitem{paszke2017automatic}
A.~Paszke, S.~Gross, S.~Chintala, G.~Chanan, E.~Yang, Z.~DeVito, Z.~Lin, A.~Desmaison, L.~Antiga, and A.~Lerer, ``Automatic differentiation in pytorch,'' in \emph{Autodiff Workshop, NIPS 2017}, 2017.

\end{thebibliography}

\begin{IEEEbiography}[{\includegraphics[width=1in,height=1.25in,clip,keepaspectratio]{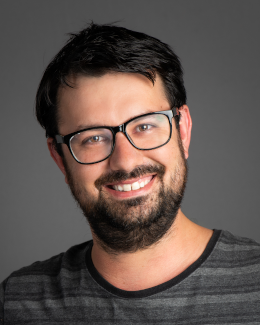}}]{Fabio A. Ruetz } 
received his B.S and M.S degrees in Mechanical and Process Engineering from the Swiss Federal Institute of Technology Zurich (ETH), Zurich, Switzerland, in 2018. Since May 2020, he has been pursuing a Ph.D. at QUT Centre for Robotics, Queensland University of Technology (QUT), in collaboration with the robotic perception group at the Commonwealth Scientific and Industrial Research Organisation (CSIRO) and Emesent. His research and interest lie in probabilistic mapping, computer vision, machine learning, and path planning to enable autonomous ground vehicles to operate in challenging environments. 
\end{IEEEbiography}

\begin{IEEEbiography}[{\includegraphics[width=1in,height=1.25in,clip,keepaspectratio]{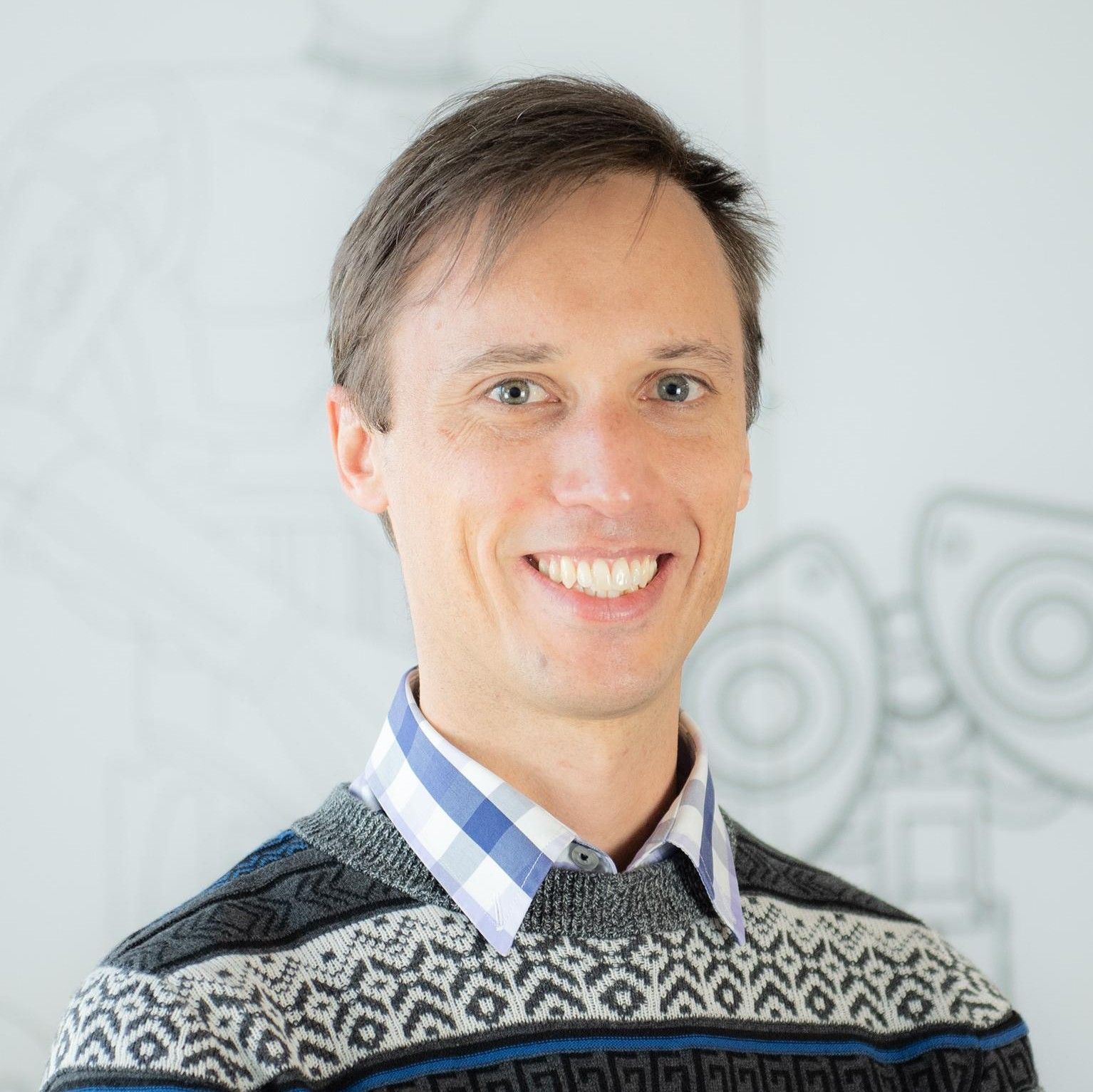}}]{Nicholas Lawrance }
 completed his PhD at the University of Sydney and worked as a postdoctoral scholar at Oregon State University, USA and ETH Zurich, Switzerland. He is currently a senior research scientist in robotic perception and autonomy at the Commonwealth Scientific and Industrial Research Organisation (CSIRO) in Australia. His research focuses on adaptive planning approaches for mobile robots, particularly in the presence of environmental uncertainty. Research interests include stochastic reasoning, adaptive sampling, and modelling of complex, uncertain phenomena. Applications include aerial, ground and underwater domains, particularly for long-duration robotic missions. Nick is a member of IEEE, an Associate Editor of IEEE Robotics and Automation Letters (RA-L), and a former Associate Editor for the International Conference on Robotics and Automation (ICRA).
\end{IEEEbiography}

\begin{IEEEbiography}[{\includegraphics[width=1in,height=1.25in,clip,keepaspectratio]{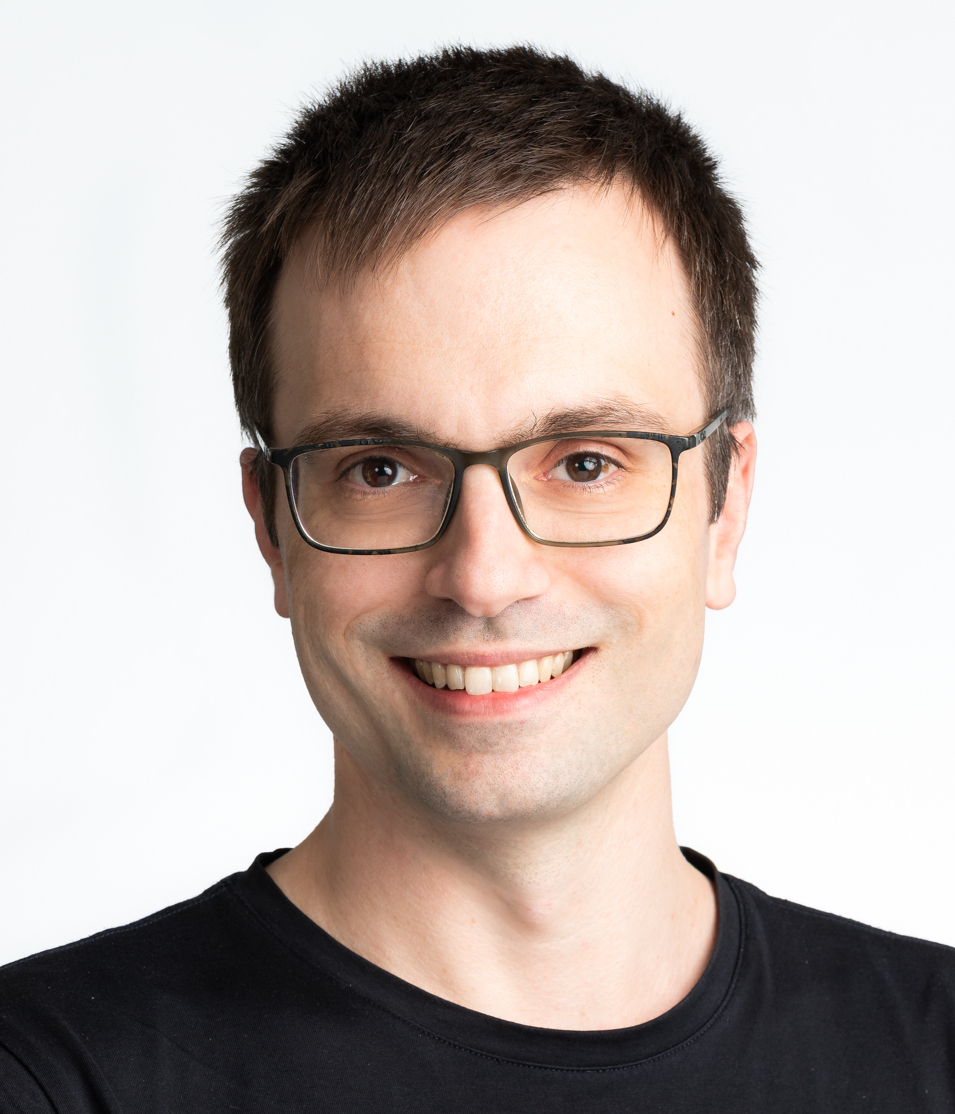}}] {Emili Hern\'andez } is an R\&D Manager with Emesent. He has two decades of experience on designing, developing and deploying novel software algorithms for underwater, ground and aerial robots. His current focus is on commercializing robotic autonomy research outcomes to improve and automate data capture in underground mining and asset inspection operations. He got his PhD at the University of Girona, Spain, and worked in several research positions at the CSIRO's Robotics and Autonomous Systems Group, Australia.
\end{IEEEbiography}

\begin{IEEEbiography}[{\includegraphics[width=1in,height=1.25in,clip,keepaspectratio]{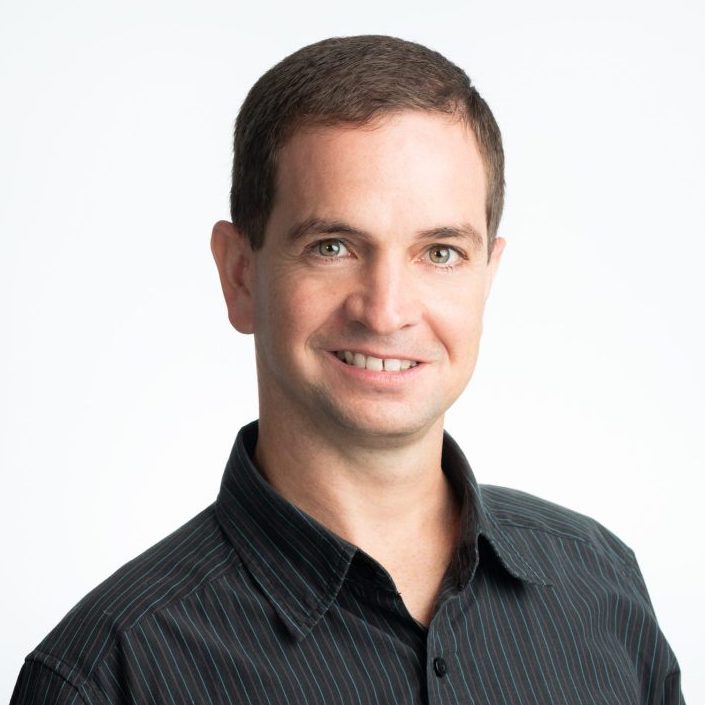}}]{Paulo V. K. Borges }
    is the Head of AI R\&D at Orica. He has a Ph.D. in Electronic Engineering and Computer Science from Queen Mary, University of London (2007). Paulo has lived and worked in different countries (USA, Brazil, UK, Switzerland, and Australia), holding positions at Orica, CSIRO, NASA, ETH Zurich, University of London, Federal University of Santa Catarina, and  University of Manchester. He holds adjunct positions as an Adjunct Associate Professor at the University of Queensland as an Adjunct Scientist at the CSIRO Data61.  His core interest has been in autonomous robots and AI solutions for the mining, manufacturing, energy, environment and space industries, with close connections between industry and research. 
\end{IEEEbiography}

\begin{IEEEbiography}[{\includegraphics[width=1in,height=1.25in,clip,keepaspectratio]{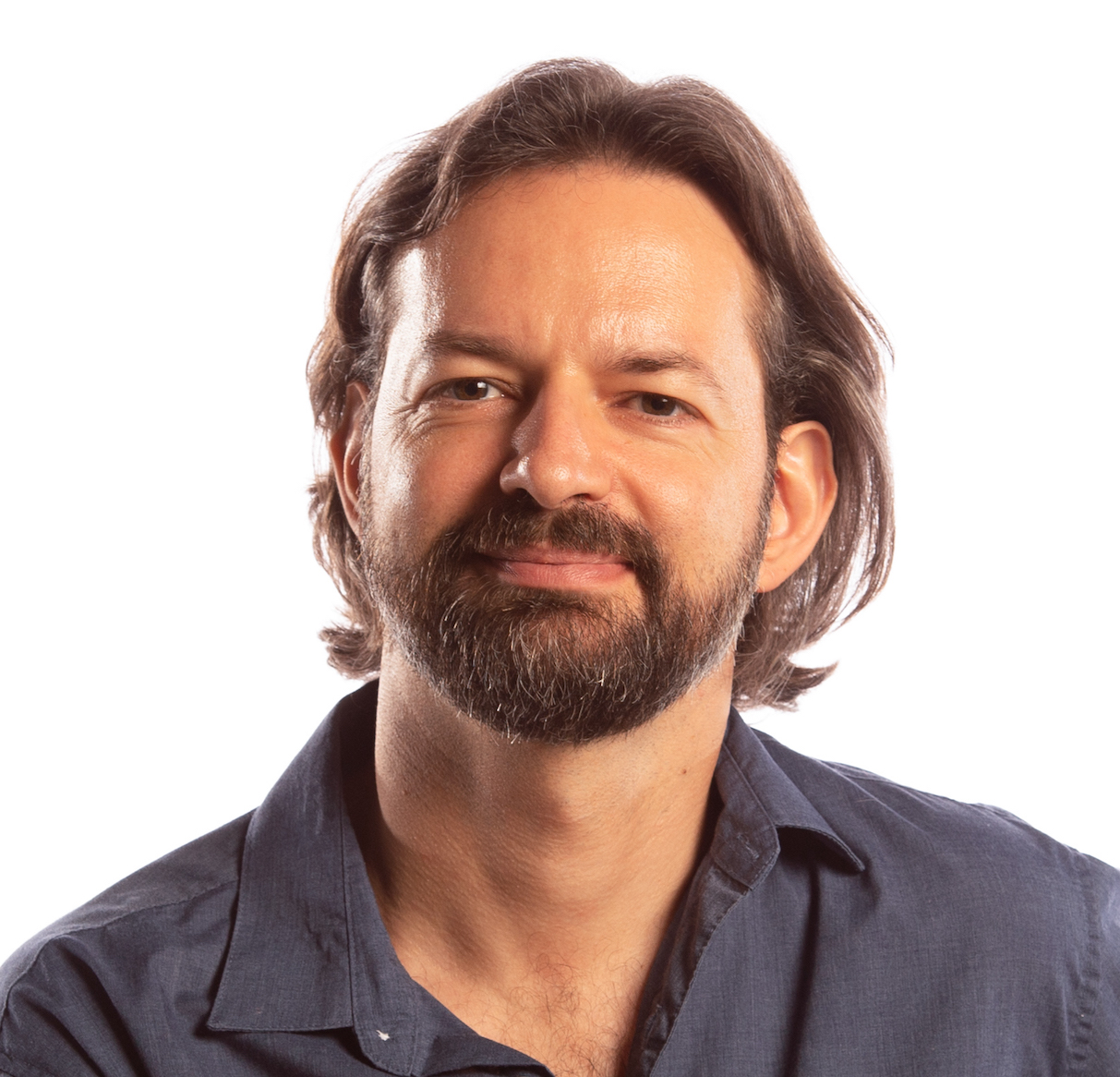}}]{Thierry Peynot }
    obtained his Ph.D. from the University of Toulouse and LAAS-CNRS in France. 
    He is Associate Professor in Robotics and Autonomous Systems at Queensland University of Technology (QUT) and a Chief Investigator of the QUT Centre for Robotics, where he leads the Mining Robotics and Space Robotics activities. Prior to joining QUT he was a researcher at the Australian Centre for Field Robotics (ACFR), The University of Sydney, worked at NASA Ames. Thierry has led multiple research programs funded by government, research institutions and industry, including mining (e.g. Caterpillar, Komatsu, Mining3), defence (e.g. BAE Systems, Rheinmetall) and space (e.g. with Boeing and CSIRO), developing robust perception technology for field robots and autonomous vehicles that can function despite adverse environmental conditions. 
    Thierry is a senior member of IEEE, immediate past Chair of the Robotics and Automation / Control Systems chapter, IEEE Queensland Section, and is a former Associate Editor of IEEE Robotics and Automation Letters (RA-L), the International Conference on Robotics and Automation (ICRA) and the International Conference on Intelligent Robots and Systems (IROS) . 
\end{IEEEbiography}

\vfill\pagebreak

\end{document}